\documentclass[10pt,twocolumn,letterpaper]{article}

\usepackage[pagenumbers]{cvpr} %

\usepackage{microtype}

\usepackage{multirow}
\usepackage{colortbl}
\definecolor{oursblue}{HTML}{2A78D6}
\definecolor{oursrow}{HTML}{EDF3FB}

\definecolor{cvprblue}{rgb}{0.21,0.49,0.74}
\usepackage[pagebackref,breaklinks,colorlinks,allcolors=cvprblue]{hyperref}

\def\confName{CVPR}
\def\confYear{2027}

\title{Structured 3D Scene Representations Improve VLM Spatial Reasoning}

\author{Simona Kocour$^{1,2,3*}$ \,\, Yutong Chen$^{3}$ \,\, Torsten Sattler$^{1}$ \,\, Sergey Prokudin$^{3}$ \,\, Siyu Tang$^{3}$\vspace*{1.5ex}\\ 
\small $^1$ Czech Institute of Informatics, Robotics and Cybernetics, Czech Technical University in Prague \\
\small $^2$ Faculty of Electrical Engineering, Czech Technical University in Prague \quad $^3$ ETH Zurich 
}

\begin{document}
\twocolumn[{%
\renewcommand\twocolumn[1][]{#1}%
\maketitle
\begin{center}
\vspace{-5pt}
\includegraphics[width=0.9\textwidth]{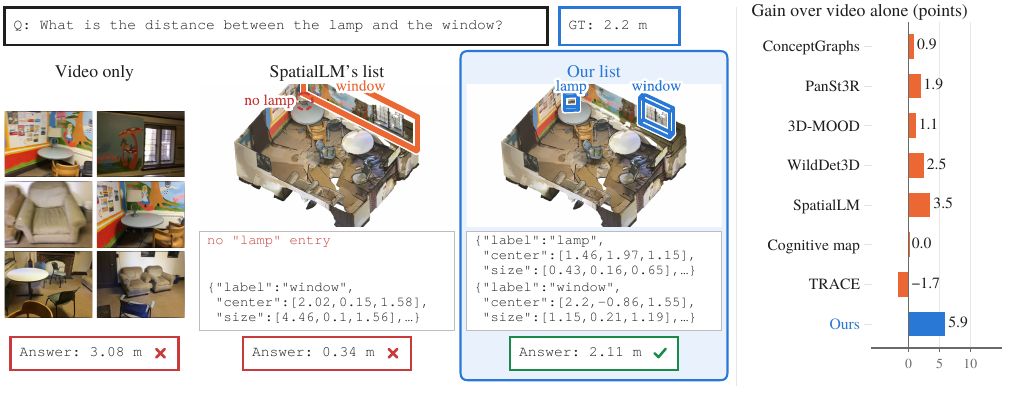}
\captionof{figure}{\textbf{The list must contain the objects a question asks about.} Left, a VSI-Bench question~\cite{yang2025thinking} answered by Gemini~3.8~Flash. From the video alone, it misjudges the distance. With the list of SpatialLM~\cite{mao2025spatiallm}, built from the same video, it answers 0.34\,m, because the list has no lamp and an enlarged window. With our list, which contains both objects with metric 3D boxes, it answers 2.11\,m, close to the true 2.2\,m. Right, the gain over the video alone on 800 VSI-Bench questions with the same VLM and frames (\cref{tab:compare}). Our list adds the most, while the lists of ConceptGraphs~\cite{gu2024conceptgraphs}, PanSt3R~\cite{zust2025panst3r}, 3D-MOOD~\cite{yang2025mood} and WildDet3D~\cite{huang2026wilddet3d} and the prompting methods cognitive map~\cite{yang2025thinking} and TRACE~\cite{hua2026trace} add little or nothing.}
\label{fig:teaser}
\end{center}
\vspace{0.5em}
}]
{\renewcommand{\thefootnote}{\fnsymbol{footnote}}\footnotetext[1]{Work done during a research visit at ETH Zurich.}}

\begin{abstract}
Vision-language models (VLMs) still struggle to reason about space from video, misjudging distances, sizes and directions. Yet simply telling them where objects are can make a large difference. Given the video together with a text list of the objects in the scene and their ground-truth metric 3D boxes, Gemini 3.8 Flash answers correctly 94.8\% of a sample of questions from the spatial benchmark VSI-Bench, compared with 68.6\% from the video alone. This 26.2-point gain shows that the VLM can use 3D geometry to answer spatial questions much more accurately than it can infer that geometry from video.
The challenge is to recover such a list from images. We therefore build an object-centric 3D representation with off-the-shelf reconstruction and segmentation models, once per scene and before any question is known. Given to the VLM alongside the original frames, it significantly improves accuracy on four spatial benchmarks, without fine-tuning any model. It also improves all five frontier VLMs we tested and provides larger gains than other 3D perception methods with the same VLM. 
In a controlled study, we introduce one type of error at a time into ground-truth lists to see which errors matter most to the VLM. Missing an object needed to answer a question costs the VLM two thirds of what the whole list adds. In contrast, adding unrelated objects barely affects its answers, although detection metrics count each of them as an error. 
Our results show that conventional detection metrics do not fully reflect what matters for VLM spatial reasoning. Covering every object a question may ask about matters more than avoiding unrelated detections.

\end{abstract}

\section{Introduction}
\label{sec:intro}
Suppose a vision-language model (VLM) watches a video of a room and is asked how far the lamp is from the window (\cref{fig:teaser}). To answer, it must find both objects and work out where they are in 3D. Current VLMs often struggle with such questions~\cite{yang2025thinking}. One approach is to fine-tune them on spatial question-answering data. Some of the strongest models on VSI-Bench are trained on questions generated from the benchmark's source datasets~\cite{yang2026cambrians,yang2026cambrianp,huang2026consispace}. Some questions can be answered without seeing the video. When those questions are removed in VSI-Bench-Debiased~\cite{brown2026debiased}, fine-tuned models lose about a tenth of their accuracy~\cite{yang2026cambrians,li2026geothinker,cai2026sensenovasi}. This suggests that they may learn shortcuts from the questions instead of relying fully on the scene.

We study a different approach. We use the VLM off the shelf, and tell it where the objects are, without fine-tuning. Gemini~3.8~Flash achieves 68.6\% average accuracy across eight spatial tasks using video alone (on a sample of 800 VSI-Bench questions~\cite{yang2025thinking}). When given the ground-truth objects and their metric 3D boxes as a text list alongside the video, its accuracy rises to 94.8\%, a gain of 26.2 points (\cref{tab:reasoners}). This is an oracle experiment, so this is an upper bound, but the result shows that VLMs can use 3D geometry to answer spatial questions much more accurately than they can infer it from video alone.

The challenge is to recover such a list from the video. 
Existing 3D perception systems reconstruct scenes and detect objects, and their outputs can be converted into the same format. Yet these representations provide much smaller gains in VLM accuracy than the ground-truth annotations (\cref{fig:teaser}, right and \cref{tab:compare}). Figure~\ref{fig:teaser} shows one example: SpatialLM~\cite{mao2025spatiallm} reconstructs the room but enlarges the window, misses the lamp, and the VLM answers the distance question incorrectly. 
These systems are generally evaluated by how accurately they reconstruct, detect or segment objects, rather than by how well their outputs support spatial reasoning. A list for a VLM has a different requirement. Since the questions are not known when it is built, it needs broad object coverage and sufficiently accurate geometry to answer questions about any part of the scene.

We build such a list from images alone, using off-the-shelf models. A feed-forward reconstruction model reconstructs metric 3D geometry and camera poses~\cite{wang2026pi3}, while an open-vocabulary segmenter finds objects across frames~\cite{carion2026sam3}. We lift the masks into 3D, merge observations of the same object across views, and fit a 3D box to each object. The list also includes the reconstructed floor area. It is built once per scene, before any question is known, and given to the VLM as text alongside the original frames.

With the same VLM and frames, our list improves accuracy by 5 to 8 percentage points on four spatial benchmarks, including two that were not used during pipeline development (\cref{tab:main}), and exceeds the best published results on four of the five benchmarks. It improves all five frontier VLMs we tested (\cref{tab:reasoners}).
The list representation complements the video. 
Under the same VLM, our list provides larger accuracy gains than the evaluated 3D perception methods~\cite{gu2024conceptgraphs, zust2025panst3r, yang2025mood,mao2025spatiallm,huang2026wilddet3d}, prompting the VLM to map~\cite{yang2025thinking} or describe the scene~\cite{hua2026trace}, or having the VLM generate its own list~\cite{openai2026gpt6astra} from the same frames. 

To understand what such a list must get right, we conduct a controlled study (\cref{sec:analysis}). Starting from VSI-Bench's ground-truth lists, we introduce one type of error at a time, such as removing, duplicating, resizing or moving an object needed to answer a question, and measure the effect on VLM accuracy. Removing that object costs two thirds of what the whole list adds. Adding objects of unrelated categories, by contrast, barely affects the VLM, although detection metrics penalize them heavily. The VLM also often follows incorrect information in the list despite having the video. 
Oracle fixes of our own lists point the same way. Of five corrections, only restoring the missed objects is needed for a significant gain.
Since the questions are unknown when the list is built, broad object coverage matters more than avoiding unrelated detections. Conventional detection metrics do not fully capture these priorities.

We make the following contributions:
\begin{itemize}
    \item We show that frontier VLMs can reason accurately about 3D space when given explicit scene geometry. On a sample of VSI-Bench, five VLMs reach 85.5--95.5\% accuracy with ground-truth object lists, without fine-tuning.
    \item We build such lists from images alone, using off-the-shelf models. Our list significantly improves spatial reasoning on four benchmarks and provides larger gains than the evaluated 3D perception and prompting methods under the same VLM.
    \item We study which errors in a 3D representation affect spatial reasoning. Our controlled experiments show that missing objects needed to answer a question is particularly costly, while conventional detection metrics often penalize errors that have little effect on the VLM.
    \item We will release the source code, the lists of all evaluated scenes, the prompts and VLM answers, together with the error-injection code and the protocol fixed before the controlled study
\end{itemize}

\section{Related Work}
\label{sec:related}

\paragraph{Spatial reasoning in VLMs.}
VSI-Bench~\cite{yang2025thinking} evaluates spatial reasoning from indoor video walkthroughs, while subsequent benchmarks cover single and multiple images~\cite{li2026spatialladder,yang2026mmsi,zhang2025spar}, mental models from sparse views~\cite{wang2026mindcube}, reasoning from another person's viewpoint~\cite{li2026viewspatial}, and situated questions in 3D scenes~\cite{ma2023sqa3d}. Some benchmarks contain questions that can be answered without visual evidence~\cite{ma2026do3dllm}, a limitation examined by VSI-Bench-Debiased~\cite{brown2026debiased}. Most approaches improve spatial reasoning by fine-tuning VLMs on spatial question-answering data~\cite{chen2024spatialvlm,cheng2024spatialrgpt,yang2026cambrians,huang2026consispace,cai2026sensenovasi,li2026spatialladder}, sometimes with 3D features~\cite{wu2025spatialmllm,fan2026vlm3r,li2026geothinker}, camera information~\cite{yang2026cambrianp}, spatial memory~\cite{liu2026vlm2,liu2026spatialttt}, or reinforcement learning~\cite{li2026starr1,chen2026spacetools}. We study a complementary setting: the VLM is off-the-shelf and receives, together with the frames, an explicit 3D representation reconstructed from them. This needs no  fine-tuning, and applies to any VLM.

\paragraph{Structured scene representations for VLMs.}
Scene graphs provide a compact description of objects and their relationships and have been used for robotic planning, navigation, and embodied question answering~\cite{rana2023sayplan,werby2024hovsg,gu2024conceptgraphs,majumdar2024openeqa}. Explicit geometry has also been used to improve spatial reasoning without fine-tuning: SandboxVLM~\cite{liu2026sandboxvlm} reasons over bouding boxes, SpatialPIN~\cite{ma2024spatialpin} uses 3D priors inferred from a single image, and Struct2D~\cite{zhu2025struct2d} supplies bird's-eye-view renderings with object annotations and metadata. Ego3D-VLM~\cite{gholami2026ego3d} builds a geometry-based cognitive map from egocentric views. Closer to our setting, GR3D~\cite{yuan2026gr3d} reconstructs object geometry from images and links textual 3D attributes to objects marked in the images. GraFT~\cite{du2026graft} operates on a 3D scene graph through geometric tools, task-specific bird's-eye views, and selected frames, evaluating both ground-truth and predicted scene graphs. We instead provide the VLM with the video frames and a wide list of objects detected in the scene, with their 3D metric bouding boxes, constructed without knowing the questions. Our experiments examine not only whether this representation improves reasoning, but also how errors in the representation affect the answers. %

\paragraph{Prompting and tool-based reasoning.}
Other training-free approaches help VLMs reason about space by marking objects across frames~\cite{liu2025coarse}, adding reconstructed views~\cite{qi2026gpt4scene}, visualizing camera motion~\cite{li2025seetrek}, or prompting the model to construct a cognitive map or describe the camera trajectory~\cite{yang2025thinking,hua2026trace}. Visual programs~\cite{gupta2023visprog,suris2023vipergpt,luo2026pyspatial} and agents~\cite{chen2026gca,marsili2025vadar,guo2026mssr} instead invoke perception or geometric tools while answering individual questions. Our pipeline performs perception once per scene, outside the VLM. The same representation is then available for every question, without question-dependent perception, rendering, or tool calls. We compare against cognitive-map prompting~\cite{yang2025thinking}, TRACE~\cite{hua2026trace}, and GCA~\cite{chen2026gca} under the same VLM.

\paragraph{3D perception and downstream evaluation.}
Object-centric 3D representations can be recovered through open-vocabulary instance segmentation~\cite{takmaz2023openmask3d,nguyen2024open3dis,yan2024maskclustering}, panoptic reconstruction~\cite{zust2025panst3r,gong2026ov3r}, open-set 3D detection~\cite{yang2025mood,zhang2025detany3d,hsu2025openm3d,cao2026vggtdet}, and scene-layout prediction~\cite{mao2025spatiallm}. These systems are generally evaluated using detection or segmentation metrics, which measure object coverage, false positives, and geometric agreement with ground truth. Prior work has analyzed detection errors separately~\cite{hoiem2012diagnosing,bolya2020tide}, while planner-centric metrics measure their effect on downstream decisions~\cite{philion2020pkl}. Language models are also known to follow incorrect textual context, including when it contradicts visual evidence~\cite{shi2023distracted,deng2025words}. 
We study how missing, duplicated and geometrically inaccurate objects in a structured 3D input affect spatial answers. By measuring these effects in a controlled setting and testing them on predicted representations, we examine whether conventional detection metrics reflect the errors that matter for VLM reasoning.

\section{A Structured 3D Representation}
\label{sec:method}

\begin{figure*}[t]
\centering
\includegraphics[width=0.9\textwidth]{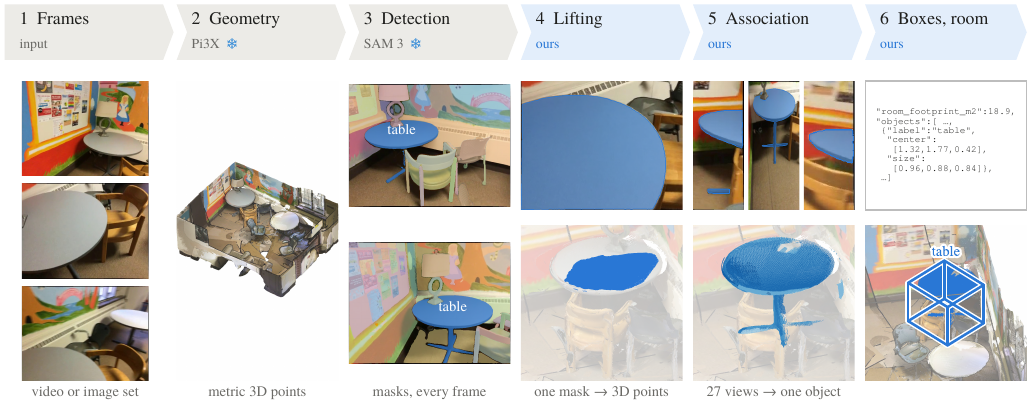}
\caption{\textbf{Building a scene-level 3D representation from images.}
Given video frames or an image set (1), Pi3X~\cite{wang2026pi3} reconstructs metric 3D points and camera poses (2). SAM~3~\cite{carion2026sam3} segments the frames, prompted with a fixed vocabulary and with object names from captions of the frames (3). We lift the masks to 3D (4), associate observations across views into objects (5), and fit metric boxes (6). The list also gives the floor area. The VLM receives the list as text, together with the frames. Snowflakes mark off-the-shelf models that we do not fine-tune.}
\vspace{-10pt}
\label{fig:pipeline}
\end{figure*}

We represent a scene as a \emph{text list} of objects, each described by a label and a metric, gravity-aligned 3D bounding box (\cref{fig:pipeline}). The list is constructed once per scene, before the questions are known. Our aim is therefore to recover every object that might be asked about, associate its observations across views without merging distinct objects, and preserve the geometry needed for spatial questions.

The pipeline takes only RGB frames, without camera intrinsics, poses or depth. For a video, we select the two sharpest frames per second, at most 106. For an image set, we use all images. Every learned component is an off-the-shelf model with released weights, and none is fine-tuned.

\subsection{Metric geometry from images}

Pi3X, the metric-scale version of $\pi^3$~\cite{wang2026pi3}, jointly estimates metric 3D points, confidence values and camera poses from the input frames. We discard pixels with a confidence below 0.6 and pixels on depth edges, where the depth varies by more than 3\% within the pixel's $3\times3$ neighborhood, the rule of $\pi^3$'s released code.
When an absolute confidence threshold would remove most points, we additionally keep those above the scene's median confidence, following the relative filtering rule of VGGT~\cite{wang2025vggt}. We estimate a gravity-aligned coordinate frame from the predicted camera orientations and fitted planes, align its horizontal axes with the dominant walls, and place the floor at $z=0$.

\subsection{Finding and grouping objects}

Because the questions are unknown, the object names can come neither from the question nor from a benchmark's labels. We use two sources of names. The first is a fixed vocabulary of 157 everyday indoor categories, which we wrote without consulting any benchmark's labels. The second comes from the scene itself. Florence-2~\cite{xiao2024florence2} writes a caption for each frame, and the 60 object names that occur most often in these captions are added to the vocabulary. SAM~3~\cite{carion2026sam3} segments every name in every frame. Non-maximum suppression across names removes overlapping masks of the same object and keeps their names as alternative labels.

We lift each mask to the reconstructed 3D points of its pixels. A point counts as a depth outlier, and is removed, if its depth differs from the median depth of the mask by more than three median absolute deviations. Observations of the same object must then be associated across views, while distinct objects of the same category must remain separate. We link two observations from different frames if their 5\,cm voxel sets overlap by at least a quarter of the smaller set and their centres are within 1.5\,m. Links are merged greedily, starting with the strongest overlap, unless a merge would join incompatible observations, for example two disjoint masks with the same name in one frame (two chairs side by side), a mask inside a differently named mask of the same frame (a pillow on a sofa), or two objects seen in several frames under names that disagree. We keep an object only if it has at least 100 points seen in two or more frames, or at least 300 points, and remove floor, wall and ceiling. Each object receives the point-weighted majority label of its masks.

\subsection{Boxes, room and interface}

For each object, we fit a 3D bounding box. For a video, we add the room's floor area, measured on the reconstruction.

The list is written as JSON text, it is a structured 3D representation. Each object has a label, a centre and three side lengths in metres. The prompt explains the coordinate frame and warns that the list may omit, duplicate or split objects. When answering a question about a video, the VLM receives 32 evenly spaced frames. For an image set, it receives all images. The same list serves every question about the scene. The VLM answers each question in a single call, from the frames and the list together, without calling perception tools or changing the list. The prompt and the pipeline settings are the same on every benchmark.

\section{Experiments}
\label{sec:experiments}

\begin{table*}[t]
\caption{\textbf{Results on five benchmarks.} Accuracy (\%) on the full test sets. The right columns give Gemini~3.8~Flash with the same frames, without and with our list. For context, the left columns give the best results published at top-tier venues, with the VLM of each method without spatial training in parentheses. They use other VLMs and serve as context only. \Cref{sec:compare} gives controlled comparisons, and \cref{sec:ablate} evaluates the list without the frames. $^\dagger$Trained on data generated from the benchmark's source datasets or with its question generator. $^\circ$Not statistically significant.}
\vspace{-2pt}
\centering
\footnotesize
\setlength{\tabcolsep}{5pt}
\begin{tabular}{@{}lllccc@{}}
\toprule
 & \multicolumn{2}{c}{Best published result} & \multicolumn{3}{c}{Gemini~3.8~Flash, same frames} \\
\cmidrule(lr){2-3}\cmidrule(l){4-6}
Benchmark & Fine-tuned for spatial reasoning & Without spatial training & Video only & + our list & Gain \\
\midrule
VSI-Bench~\cite{yang2025thinking} & 76.6$^\dagger$ ConsiSpace~\cite{huang2026consispace} & 60.2 TRACE~\cite{hua2026trace} (Gemini-3-Pro) & 67.5 & \textbf{73.6} & 6.0 \\
VSI-Bench-Debiased~\cite{brown2026debiased} & 66.3 GeoThinker~\cite{li2026geothinker} & -- & 65.3 & \textbf{71.6} & 6.3 \\
MMSI-Bench~\cite{yang2026mmsi} & 43.3 SenseNova-SI~\cite{cai2026sensenovasi} & 55.0 GCA~\cite{chen2026gca} (Gemini-2.5-Pro) & 64.1 & \textbf{72.2} & 8.1 \\
SPBench-MV~\cite{li2026spatialladder} & 70.9$^\dagger$ SpatialLadder~\cite{li2026spatialladder} & 61.9 GCA~\cite{chen2026gca} (Qwen3-VL) & 72.8 & \textbf{78.0} & 5.2 \\
MindCube-Tiny~\cite{wang2026mindcube} & 85.7$^\dagger$ SenseNova-SI~\cite{cai2026sensenovasi} & 64.2 GCA~\cite{chen2026gca} (Qwen3-VL) & 92.4 & \textbf{94.3} & 1.9$^\circ$ \\
\bottomrule
\end{tabular}
\label{tab:main}
\end{table*}

\subsection{Setup}
\label{sec:setup}

We evaluate on VSI-Bench~\cite{yang2025thinking}, 5130 questions about 288 videos of indoor rooms from ScanNet~\cite{dai2017scannet}, ScanNet++~\cite{yeshwanth2023scannetpp} and ARKitScenes~\cite{baruch2021arkitscenes}, and on VSI-Bench-Debiased~\cite{brown2026debiased}, a subset of 2362 of its questions. We used VSI-Bench while developing the pipeline. Three further benchmarks were not used during development, MMSI-Bench~\cite{yang2026mmsi} (1000 questions about sets of images), SPBench-MV~\cite{li2026spatialladder} (319) and MindCube-Tiny~\cite{wang2026mindcube} (1050).

Accuracy is the percentage of correct answers. For numerical answers such as distances, VSI-Bench and SPBench-MV give partial credit by how close the answer is, averaged over several tolerances (mean relative accuracy~\cite{yang2025thinking}), and both benchmarks average the accuracy over their tasks. A gain is a difference in accuracy, in percentage points. It is significant if its 95\% confidence interval excludes zero, where the interval comes from a paired bootstrap that resamples scenes, or image sets for the image benchmarks~\cite{efron1993bootstrap}. Tables mark gains that are not significant with $^\circ$.

All results use Gemini~3.8~Flash~\cite{google2025gemini3} unless another VLM is named. \Cref{tab:reasoners} also tests Gemini~3.5~Flash-Lite and 3.1~Pro~\cite{comanici2025gemini}, GPT-5~\cite{singh2025gpt5}, GPT-6 Astra~\cite{openai2026gpt6astra}, and the open-weight Qwen3-VL-8B and 32B~\cite{bai2025qwen3vl} (settings in the supplement). In every comparison, the VLM answers the same question with the same visual input, 32 frames of a video or all images of an image set, once without and once with a list, and only the list differs. The \emph{ground-truth list} is built in our format from the benchmark's own 3D annotation, from which its answers were generated, and serves as an upper bound.
\Cref{tab:main,tab:pertask} use the full test sets. All other VSI-Bench results use the same 800 questions, 100 per task, drawn at random before the experiments (\cref{tab:reasoners}).
The hypotheses and statistical tests for the controlled error study (\cref{sec:analysis}) were specified before the experiments. 

\subsection{Results across benchmarks and VLMs}
\label{sec:results}

\begin{figure*}[t]
\centering
\includegraphics[width=0.95\textwidth]{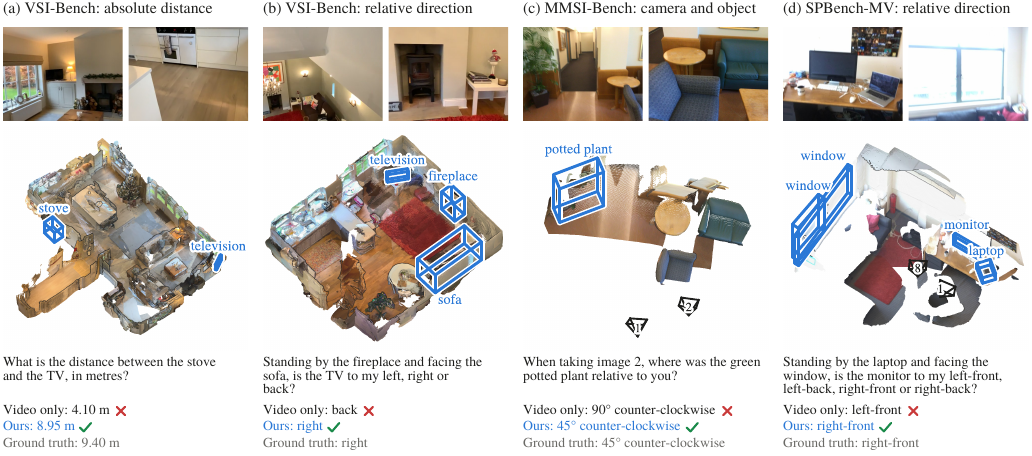}
\caption{\textbf{Examples across benchmarks} (Gemini~3.8~Flash). Each panel shows two input images and our 3D representation visualized on the reconstructed scene. Objects referred to by the question are highlighted in blue; camera poses are also shown for image sets. In these selected examples, the VLM answers incorrectly from the images alone and correctly with our representation. Questions are shortened for display.}
\label{fig:qual}
\end{figure*}

\Cref{tab:main} shows the results. With Gemini~3.8~Flash and the same frames, our list raises accuracy on VSI-Bench from 67.5\% to 73.6\%. The gain of 6.3 points on VSI-Bench-Debiased shows that it does not come from questions that can be answered without looking, whereas fine-tuned models lose 6 to 8 points there~\cite{yang2026cambrians,cai2026sensenovasi,li2026geothinker}. On the image benchmarks that we did not use during development, the unchanged pipeline adds 8.1 points on MMSI-Bench and 5.2 on SPBench-MV. On MindCube-Tiny, the images alone already give 92.4\%, which leaves little room, and the gain of 1.9 points is not significant. \Cref{fig:qual} shows examples. With our list, Gemini~3.8~Flash is more accurate than every published result in \cref{tab:main} but one, with or without spatial training. The exception is ConsiSpace~\cite{huang2026consispace} on VSI-Bench, which is fine-tuned on questions generated in the style of VSI-Bench from the same source datasets. These results use other VLMs, so \cref{sec:compare} compares representations with the same VLM, and our list is again the most accurate.

\begin{table}[t]
\caption{\textbf{Accuracy by task on VSI-Bench.} All 5,130 questions, Gemini~3.8~Flash with the same frames, without and with our list. Tasks that geometry decides gain most. Object size loses because boxes can cover only part of a large object. $^\circ$Not statistically significant.}
\vspace{-5pt}
\centering
\footnotesize
\setlength{\tabcolsep}{4pt}
\begin{tabular}{@{}lccc@{}}
\toprule
Task & Video & + list & Gain \\
\midrule
Room size & 54.2 & 75.2 & 21.0 \\
Abs.\ distance & 52.2 & 69.7 & 17.5 \\
Route plan & 79.9 & 91.8 & 11.9 \\
Rel.\ direction & 76.4 & 84.4 & 8.0 \\
Appear.\ order & 73.0 & 74.3 & 1.3$^\circ$ \\
Counting & 56.9 & 56.0 & \ensuremath{-}0.8$^\circ$ \\
Rel.\ distance & 71.3 & 68.2 & \ensuremath{-}3.1$^\circ$ \\
Object size & 76.5 & 69.1 & \ensuremath{-}7.4 \\
\midrule
All (task average) & 67.5 & 73.6 & 6.0 \\
\bottomrule
\end{tabular}
\vspace{-10pt}
\label{tab:pertask}
\end{table}

The gains follow what the list states explicitly (\cref{tab:pertask}). Room size gains most, because our list gives the floor area of the room, which our pipeline measures. Absolute distance, route planning and relative direction gain because the list places all objects in one metric frame, so the VLM can compute distances and directions instead of estimating them across views. Appearance order, counting and relative distance do not change significantly. The list does not record when an object is seen, and its potential missed, duplicated or partial boxes can mislead counts and comparisons. Object size gets worse because the box of a large object can cover only the part that was reconstructed and segmented, and the VLM reports the size of the box instead of correcting it from the video, as it often follows the list over the frames (\cref{sec:analysis}).

The list does not replace the video. Given only the list, without any frame, Gemini~3.8~Flash is about 6 points less accurate on the 800 questions than from the video alone, and 11.8 points less accurate than with both (\cref{tab:ablate}). The list alone is more accurate than the video alone on tasks that geometry decides, such as absolute distance (60.1\% \vs 48.8\%) and relative direction (83.0\% \vs 72.0\%). It fails on appearance order (39.0\% \vs 76.0\%), because the list does not record when an object is seen, and on counting (32.6\% \vs 62.3\%), where the frames correct objects that the list misses or duplicates. In all other results, the VLM therefore receives the list together with the frames.

\Cref{tab:reasoners} repeats the comparison with other VLMs. On the 800 questions, our list raises the accuracy of all three Gemini models~\cite{comanici2025gemini} significantly, by 5.9 to 10.2 points. our list adds about 10 points to GPT-5~\cite{singh2025gpt5} and GPT-6 Astra~\cite{openai2026gpt6astra}. With the ground-truth list, the Gemini and GPT models reach 85.5\% to 95.5\%, so better 3D perception would pay off further. The open-weight Qwen3-VL models~\cite{bai2025qwen3vl} do not gain from our list. Both gain significantly from the ground-truth list, so they can use the format, but they are not robust to the errors in our list. The list also makes answers cheaper. With it, Gemini~3.8~Flash uses less than half as many thinking tokens per question, so an answer costs about 40\% less although the prompt is longer (see supplement).

\begin{table}[t]
\caption{\textbf{Our list helps every frontier VLM.} Accuracy (\%) with the same 32 frames for every VLM. The 800 VSI-Bench questions of \cref{tab:compare}, 100 per task. Each gain is over the video alone, computed before rounding. The ground-truth (GT) list is an upper bound. $^\ddagger$Open-weight models. $^\circ$Not statistically significant.}
\centering
\footnotesize
\setlength{\tabcolsep}{3pt}
\begin{tabular}{@{}lccccc@{}}
\toprule
VLM & Video & + our list & Gain & + GT list & Gain \\
\midrule
\multicolumn{6}{@{}l}{\textit{800 questions, 100 per task}} \\
Gemini~3.5~Flash-Lite & 53.6 & 61.5 & 7.9 & 85.5 & 31.9 \\
Gemini~3.1~Pro & 59.4 & 69.6 & 10.2 & 91.4 & 32.0 \\
Gemini~3.8~Flash & 68.6 & 74.5 & 5.9 & 94.8 & 26.2 \\
GPT-5 & 51.2 & 61.4 & 10.1 & 87.9 & 36.7 \\
GPT-6 Astra & 65.0 & 74.7 & 9.7 & 95.5 & 30.4 \\
Qwen3-VL-8B-Thinking$^\ddagger$ & 51.4 & 51.9 & 0.5$^\circ$ & 68.2 & 16.8 \\
Qwen3-VL-32B-Instruct$^\ddagger$ & 63.2 & 61.8 & \ensuremath{-}1.4$^\circ$ & 70.7 & 7.5 \\
\bottomrule
\end{tabular}
\label{tab:reasoners}
\end{table}

\subsection{Comparison with the same VLM}
\label{sec:compare}
\begin{table}[t]
\caption{\textbf{Comparison with other methods.} Gemini~3.8~Flash with the same frames on 800 VSI-Bench questions, 100 per task. Accuracy (\%) and gain over the video alone. The row \emph{ConceptGraphs, our input} applies its association step to our reconstruction and masks. $^\circ$Not statistically significant.}
\centering
\footnotesize
\setlength{\tabcolsep}{3pt}
\begin{tabular}{@{}lcc@{}}
\toprule
Method & Accuracy & Gain \\
\midrule
Video only & 68.6 & -- \\
\midrule
Cognitive map~\cite{yang2025thinking} & 68.6 & 0.0$^\circ$ \\
TRACE~\cite{hua2026trace} & 66.9 & \ensuremath{-}1.7$^\circ$ \\
\midrule
ConceptGraphs~\cite{gu2024conceptgraphs} & 69.4 & 0.9$^\circ$ \\
PanSt3R~\cite{zust2025panst3r} & 70.5 & 1.9$^\circ$ \\
3D-MOOD~\cite{yang2025mood} & 69.7 & 1.1$^\circ$ \\
3D-MOOD, fused across frames & 68.4 & \ensuremath{-}0.1$^\circ$ \\
WildDet3D~\cite{huang2026wilddet3d} & 71.0 & 2.5$^\circ$ \\
SpatialLM~\cite{mao2025spatiallm} & 72.1 & 3.5 \\
\midrule
ConceptGraphs, our input & 73.6 & 5.0 \\
\textbf{Our list} & \textbf{74.5} & \textbf{5.9} \\
Ground-truth list & 94.8 & 26.2 \\
\bottomrule
\end{tabular}
\label{tab:compare}
\end{table}

We compare our list with the representations of five published 3D perception systems, converted into our format and gravity-aligned frame, so that the same VLM answers the same 800 questions with the same frames (\cref{tab:compare}). They are the robot scene graph ConceptGraphs~\cite{gu2024conceptgraphs}, the panoptic reconstruction PanSt3R~\cite{zust2025panst3r}, the single-image open-set 3D detectors 3D-MOOD~\cite{yang2025mood} and WildDet3D~\cite{huang2026wilddet3d}, whose per-frame boxes we collect into one list, and the layout model SpatialLM~\cite{mao2025spatiallm}, which we run on our reconstruction. For 3D-MOOD, we also fuse the boxes of one object across frames. PanSt3R, 3D-MOOD and, on ARKitScenes, ConceptGraphs receive the datasets' ground-truth camera intrinsics, which our pipeline does not use. Where a system gives no list for a scene, the VLM answers from the video alone, and the result is the same on the questions each system covers (see supplement). ConceptGraphs was run for an earlier version of this work, on our reconstruction and masks of that time, and adds only 0.9 points over the video alone. Applied to our final reconstruction and masks, its association step is about as accurate as our full pipeline (73.6\% \vs 74.5\%), so the gain comes mainly from our reconstruction and masks.

The lists of the published systems add at most 3.5 points over the video alone, against 5.9 for ours, and only SpatialLM's gain is significant. Two properties of the lists explain the difference. The first is metric geometry. On absolute distance, our list adds 28.9 points and the other lists 2.8 to 9.8. The second is whether the list contains the objects a question asks about, since a missing object costs the VLM two thirds of what the whole list adds (\cref{sec:analysis}).

Asking the VLM to recover the layout itself does not help. Drawing a cognitive map first~\cite{yang2025thinking} changes accuracy by 0.0 points, and TRACE~\cite{hua2026trace}, which has the VLM describe the scene and its camera path, by $-$1.7, neither significantly. 
Nor does a list that the VLM writes itself help. Asked to write a list in our format from the same up to 106 frames, once per scene and without any question, GPT-6 Astra~\cite{openai2026gpt6astra} lists about 120 objects per scene. It then answers no better than from the video alone and 10.5 points worse than with our list.
All three methods ask the VLM to infer the layout from the frames, so they inherit the errors of its perception, whereas our list adds geometry measured by a reconstruction model. The tool-using agent GCA~\cite{chen2026gca} calls reconstruction, detection and code tools for every question. With Gemini~3.8~Flash, and GroundingDINO~\cite{liu2024groundingdino} in place of its own detector, which supports only two specific VLMs, it reaches 62.8\% on MMSI-Bench, below the images alone (64.1\%) and our list (72.2\%).

\subsection{Ablations}
\label{sec:ablate}

\begin{table}[t]
\caption{\textbf{Ablations on VSI-Bench.} Change in accuracy, in percentage points, when one part of our list is changed, with Gemini~3.8~Flash and the same frames, on the 800 questions of \cref{tab:reasoners,tab:compare}. With the full list, accuracy is 74.5\%. $^\circ$Not statistically significant.}
\centering
\footnotesize
\setlength{\tabcolsep}{3pt}
\begin{tabular}{@{}lc@{}}
\toprule
Change to our list & Accuracy change \\
\midrule
Names only, without positions or sizes & \ensuremath{-}8.6 \\
Only names that map to VSI-Bench categories & \ensuremath{-}2.6 \\
Half of the objects removed at random & \ensuremath{-}3.8 \\
Fixed vocabulary only, no caption names & \ensuremath{-}0.5$^\circ$ \\
Stricter detection threshold & \ensuremath{-}0.5$^\circ$ \\
Only objects seen in at least three frames & \ensuremath{-}0.2$^\circ$ \\
List only, without the frames (text only) & \ensuremath{-}11.8 \\
No list (video alone) & \ensuremath{-}5.9 \\
\bottomrule
\end{tabular}
\label{tab:ablate}
\end{table}

\Cref{tab:ablate} removes one part of the list at a time. Names without positions and sizes do not help. They are 8.6 points less accurate than the full list and even below the video alone, so the gain comes from the geometry. Coverage matters as well. Removing half of the objects at random costs 3.8 points. Keeping only the objects whose names map to one of VSI-Bench's categories costs 2.6 points, although the questions ask only about these categories. Our names are open-vocabulary, so an asked-about object can carry a name that is similar to its category but not identical, and the mapping then drops it.

Variants that make the list more precise, a stricter detection threshold or only objects seen in at least three frames, change nothing measurable. This agrees with \cref{sec:analysis}, where extra unrelated objects barely affect the VLM. Our object names come from two sources, a fixed vocabulary and the captions of the frames (\cref{sec:method}). Without the caption names, accuracy on VSI-Bench stays the same, because almost every object category of VSI-Bench is already in the fixed vocabulary. The captions add names for objects outside this vocabulary, which other scenes may contain. The last two rows of \cref{tab:ablate}, the list without the frames and the frames without the list, are discussed in \cref{sec:results}.

The list therefore needs metric geometry and the objects that questions ask about, while a more precise list with fewer objects does not help.

\section{What Does a VLM Need from a 3D Representation?}
\label{sec:analysis}

Our representation improves spatial reasoning more than the evaluated 3D perception systems, although all are lists of objects with 3D boxes in the same format. We investigate which errors in these lists affect the VLM's answers. We first introduce controlled errors into ground-truth representations, then examine whether the findings hold for predicted lists.
\begin{table}[t]
\caption{\textbf{What one error costs the VLM and the detection metrics.} Each row changes VSI-Bench's ground-truth list in one way, on the 592 questions that ask about specific objects (572 for unrelated objects). The VLM columns give the accuracy Gemini~3.8~Flash loses against the intact list, over all six tasks, the three numerical tasks (counting, absolute distance, object size) and the three multiple-choice tasks (relative distance, relative direction, route planning). \emph{All} is the mean of \emph{Num.} and \emph{Choice}. \emph{Match} is the share of changed boxes that still count as correct detections at 3D IoU 0.25. The last two columns give the points lost by F1 at 3D IoU 0.25 and by the open detection score~\cite{yang2025mood}, both computed over all objects of a scene. $^\circ$Not statistically significant.}
\centering
\footnotesize
\setlength{\tabcolsep}{2.2pt}
\begin{tabular}{@{}lcccccc@{}}
\toprule
 & \multicolumn{3}{c}{Lost by the VLM} & Match & \multicolumn{2}{c}{Score lost} \\
\cmidrule(lr){2-4}\cmidrule(l){6-7}
Change to the GT list & All & Num. & Choice & @.25 & F1@.25 & ODS \\
\midrule
\multicolumn{7}{@{}l}{\textit{The object the question asks about}} \\
~~Missing & 17.7 & 29.3 & 6.2 & 0 & 9.1 & 5.0 \\
~~Duplicated & 6.3 & 11.6 & 1.1$^\circ$ & 100 & 6.0 & 3.0 \\
~~Size $\times$2 & 23.9 & 45.7 & 2.1$^\circ$ & 0 & 13.5 & 2.0 \\
~~Shifted 1\,m & 14.5 & 18.6 & 10.4 & 4 & 12.6 & 6.2 \\
~~Shifted 0.5\,m & 7.6 & 11.9 & 3.3 & 16 & 9.8 & 4.0 \\
\multicolumn{7}{@{}l}{\textit{The rest of the scene}} \\
~~Unrelated objects added & 1.7 & 1.5 & 1.9$^\circ$ & -- & 20.2 & 10.1 \\
\midrule
No list & 26.2 & 36.4 & 15.9 & -- & 100.0 & 100.0 \\
\bottomrule
\end{tabular}
\label{tab:errors}
\end{table}

\subsection{Controlled errors in ground-truth lists}
\label{sec:study}

We start from VSI-Bench's ground-truth lists, built from the annotation from which its answers were generated. They give each object's name, the position of its centre and its size in metres, and the room's floor area. We change one thing at a time, following error analyses of object detectors~\cite{hoiem2012diagnosing,bolya2020tide,caesar2020nuscenes}. The object a question asks about is removed, duplicated, doubled in size, or shifted by 0.5\,m or 1\,m, or objects of categories that no question asks about are added to the scene. The supplement defines each change.

Gemini~3.8~Flash answers every question with the same 32 frames and the same prompt, which warns that the list may contain errors. We leave out two of VSI-Bench's eight tasks. Room size asks about no object, and appearance order is answered from the video, because the list does not record when an object is seen. This leaves 592 of the 800 questions. For each change, we report how much accuracy the VLM loses compared with the intact ground-truth list.

We reuse the changed lists of a larger study, reported in full in the supplement, answering 38,582 queries with more kinds of errors. Before that study began, we wrote down nine hypotheses, such as that a missing object costs more than a duplicated one, together with how each would be tested. 

\subsection{Which errors matter?}
\label{sec:metrics}

Missing the object a question asks about is among the costliest errors (\cref{tab:errors}). It costs 17.7 points, two thirds of what the whole list adds (26.2), because a question concerns one or two specific objects and without them the list cannot answer it. Adding objects of unrelated categories costs only 1.7 points. The loss from a missing object is largest on numerical questions (29.3 points), where the VLM takes the count, distance or size from the list, and smaller on multiple-choice questions (6.2), where it can fall back on the video.

Geometric errors matter mainly for numerical answers. A box doubled in size costs 45.7 points on numerical questions and nothing significant on multiple-choice ones. A shift by 0.5\,m or 1\,m costs 7.6 or 14.5 points over the six tasks, more on numerical than on multiple-choice questions.

The VLM often follows a wrong list even though it sees the video. For errors whose effect on the answer has a known direction, such as a doubled box on a size question or a duplicated object on a counting question, it moves its answer in that direction 66.4\% of the time. Its answers from the video alone move that way only 21.5\% of the time under the same rule. This resembles how VLMs follow text that contradicts the image~\cite{deng2025words}.

Detection metrics weigh these errors differently. We compute the two box metrics used to evaluate SpatialLM~\cite{mao2025spatiallm} and 3D-MOOD~\cite{yang2025mood}. The first is F1 at 3D IoU 0.25, where a box counts as a correct detection if it has the right name and its intersection over union with a ground-truth box is at least 0.25. The second is the open detection score (ODS)~\cite{yang2025mood}, which also penalizes errors in centre, size and orientation. We use their F1-based form, because the lists carry no confidence scores. Unlike the VLM's answers, both metrics average over all objects of a scene. Unrelated extra objects cost F1 20.2 points and the VLM only 1.7. A 1\,m shift costs F1 more than a missing object (12.6 \vs 9.1 points), whereas the VLM loses more from the missing object (17.7 \vs 14.5). A doubled box, the costliest error for the VLM, lowers ODS by only 2.0 points.

The same mismatch appears in real lists. ConceptGraphs' list has about the same F1@0.25 as ours (0.33 \vs 0.32) and a higher precision (0.50 \vs 0.27, see supplement). Yet it contains the asked-about objects far less often, and it adds less to the VLM's accuracy (\cref{tab:compare}). 
Oracle fixes of an earlier version of our lists point the same way. We correct five kinds of errors with ground truth and leave out one correction at a time. Only leaving out the restored missing objects costs significant accuracy (see supplement).
Detection scores alone therefore do not tell whether a list helps a VLM, because what matters depends on which objects and measurements a question needs (\cref{tab:errors}).

\section{Discussion}
\label{sec:discussion}

Because the question is unknown when a 3D representation for a VLM is built, it should first of all contain every object that may be asked about, while extra objects of unrelated categories cost little. Precision still matters for the asked-about objects, since a duplicate misleads counting and a wrong box misleads numerical answers. The ground-truth lists show how much better perception could still help. For most questions, our list still misses at least one of the objects the question asks about.

Our study has limitations. The list describes a static scene, so on MMSI-Bench questions~\cite{yang2026mmsi} about camera motion, accuracy drops from 83.8\% to 71.6\% with it. Boxes of large objects can be incomplete, which lowers accuracy on object size. The open-weight VLMs we tested do not gain from our list, although they gain from the ground-truth list. These smaller models may need fine-tuning to cope with the errors in the list, which we leave to future work.

\section{Conclusion}
\label{sec:conclusion}

VLMs answer spatial questions far more accurately when they are told where the objects are. We build such a list of objects and their metric 3D boxes from images alone, with off-the-shelf models, once per scene and before any question is known. Given to the VLM together with the frames, it improves accuracy on four spatial benchmarks and for all five frontier VLMs we tested, without fine-tuning, and it helps more than the outputs of other 3D perception systems. Our controlled study shows what such a list must get right. A missing asked-about object costs two thirds of what the whole list adds, whereas unrelated extra objects barely matter, although detection metrics count them as errors. Because VLMs also follow lists that are wrong, 3D perception for VLMs should be judged by the questions it lets them answer, and not by detection metrics alone.

\section*{Acknowledgments}
This work was supported by 
the Czech Science Foundation (GACR) EXPRO (grant no. 23-07973X),
the Ministry of Education, Youth and Sports of the Czech Republic through the e-INFRA CZ (ID:90254),
the Grant Agency of the Czech Technical University in Prague, grant No. SGS26/098/OHK3/1T/13.
Yutong Chen is funded by a Google PhD Fellowship.
The author(s) used generative AI to assist with editing the manuscript, cleaning up figures and coding. All was reviewed by the authors, who take full responsibility for its content.

{
    \small
    \bibliographystyle{ieeenat_fullname}
    \bibliography{main}

\begin{thebibliography}{79}
\providecommand{\natexlab}[1]{#1}
\providecommand{\url}[1]{\texttt{#1}}
\expandafter\ifx\csname urlstyle\endcsname\relax
  \providecommand{\doi}[1]{doi: #1}\else
  \providecommand{\doi}{doi: \begingroup \urlstyle{rm}\Url}\fi

\bibitem[Bai et~al.(2025)]{bai2025qwen3vl}
Shuai Bai et~al.
\newblock {Qwen3-VL} technical report.
\newblock \emph{arXiv preprint arXiv:2511.21631}, 2025.

\bibitem[Baruch et~al.(2021)Baruch, Chen, Dehghan, Dimry, Feigin, Fu, Gebauer,
  Joffe, Kurz, Schwartz, and Shulman]{baruch2021arkitscenes}
Gilad Baruch, Zhuoyuan Chen, Afshin Dehghan, Tal Dimry, Yuri Feigin, Peter Fu,
  Thomas Gebauer, Brandon Joffe, Daniel Kurz, Arik Schwartz, and Elad Shulman.
\newblock {ARKitScenes}: A diverse real-world dataset for {3D} indoor scene
  understanding using mobile {RGB-D} data.
\newblock In \emph{Advances in Neural Information Processing Systems (NeurIPS),
  Datasets and Benchmarks Track}, 2021.

\bibitem[Bolya et~al.(2020)Bolya, Foley, Hays, and Hoffman]{bolya2020tide}
Daniel Bolya, Sean Foley, James Hays, and Judy Hoffman.
\newblock {TIDE}: A general toolbox for identifying object detection errors.
\newblock In \emph{Computer Vision -- {ECCV} 2020}, pages 558--573. Springer,
  2020.

\bibitem[Brown et~al.(2026)Brown, Yang, Yang, Fergus, and
  Xie]{brown2026debiased}
Ellis Brown, Jihan Yang, Shusheng Yang, Rob Fergus, and Saining Xie.
\newblock Benchmark designers should ``train on the test set'' to expose
  exploitable non-visual shortcuts.
\newblock In \emph{Conference on Language Modeling (COLM)}, 2026.

\bibitem[Caesar et~al.(2020)Caesar, Bankiti, Lang, Vora, Liong, Xu, Krishnan,
  Pan, Baldan, and Beijbom]{caesar2020nuscenes}
Holger Caesar, Varun Bankiti, Alex~H. Lang, Sourabh Vora, Venice~Erin Liong,
  Qiang Xu, Anush Krishnan, Yu Pan, Giancarlo Baldan, and Oscar Beijbom.
\newblock {nuScenes}: A multimodal dataset for autonomous driving.
\newblock In \emph{IEEE/CVF Conference on Computer Vision and Pattern
  Recognition (CVPR)}, pages 11621--11631, 2020.

\bibitem[Cai et~al.(2026)Cai, Wang, Gu, Pu, Xu, Wang, Yin, Yang, Wei, Zhou,
  Sun, Pang, Li, Qian, Lin, Shi, Deng, Han, Chen, Fan, Deng, Lu, Pan, Li, Liu,
  Wang, Lin, and Yang]{cai2026sensenovasi}
Zhongang Cai, Ruisi Wang, Chenyang Gu, Fanyi Pu, Junxiang Xu, Yubo Wang, Wanqi
  Yin, Zhitao Yang, Chen Wei, Tongxi Zhou, Qingping Sun, Hui~En Pang, Jiaqi Li,
  Oscar Qian, Zhiqian Lin, Xuanke Shi, Kewang Deng, Xiaoyang Han, Zukai Chen,
  Xiangyu Fan, Hanming Deng, Lewei Lu, Liang Pan, Bo Li, Ziwei Liu, Quan Wang,
  Dahua Lin, and Lei Yang.
\newblock Scaling spatial intelligence with multimodal foundation models.
\newblock In \emph{Proceedings of the IEEE/CVF Conference on Computer Vision
  and Pattern Recognition (CVPR)}, pages 7879--7890, 2026.

\bibitem[Cao et~al.(2026)Cao, Wu, Chen, Zhong, Hong, and Xu]{cao2026vggtdet}
Yang Cao, Feize Wu, Dave~Zhenyu Chen, Yingji Zhong, Lanqing Hong, and Dan Xu.
\newblock {VGGT-Det}: Mining {VGGT} internal priors for sensor-geometry-free
  multi-view indoor {3D} object detection.
\newblock In \emph{Proceedings of the IEEE/CVF Conference on Computer Vision
  and Pattern Recognition (CVPR)}, pages 4708--4717, 2026.

\bibitem[Carion et~al.(2026)Carion, Gustafson, Hu, Debnath, Hu, Suris
  Coll-Vinent, Ryali, Alwala, Khedr, Huang, Lei, Ma, Guo, Kalla, Marks, Greer,
  Wang, Sun, R{\"a}dle, Afouras, Mavroudi, Xu, Wu, Zhou, Momeni, Hazra, Ding,
  Vaze, Porcher, Li, Li, Kamath, Cheng, Doll{\'a}r, Ravi, Saenko, Zhang, and
  Feichtenhofer]{carion2026sam3}
Nicolas Carion, Laura Gustafson, Yuan-Ting Hu, Shoubhik Debnath, Ronghang Hu,
  Didac Suris Coll-Vinent, Chaitanya Ryali, Kalyan~Vasudev Alwala, Haitham
  Khedr, Andrew Huang, Jie Lei, Tengyu Ma, Baishan Guo, Arpit Kalla, Markus
  Marks, Joseph Greer, Meng Wang, Peize Sun, Roman R{\"a}dle, Triantafyllos
  Afouras, Effrosyni Mavroudi, Katherine Xu, Tsung-Han Wu, Yu Zhou, Liliane
  Momeni, Rishi Hazra, Shuangrui Ding, Sagar Vaze, Francois Porcher, Feng Li,
  Siyuan Li, Aishwarya Kamath, Ho~Kei Cheng, Piotr Doll{\'a}r, Nikhila Ravi,
  Kate Saenko, Pengchuan Zhang, and Christoph Feichtenhofer.
\newblock {SAM} 3: Segment anything with concepts.
\newblock In \emph{International Conference on Learning Representations
  (ICLR)}, 2026.

\bibitem[Chen et~al.(2024)Chen, Xu, Kirmani, Ichter, Sadigh, Guibas, and
  Xia]{chen2024spatialvlm}
Boyuan Chen, Zhuo Xu, Sean Kirmani, Brian Ichter, Dorsa Sadigh, Leonidas
  Guibas, and Fei Xia.
\newblock {SpatialVLM}: Endowing vision-language models with spatial reasoning
  capabilities.
\newblock In \emph{Proceedings of the IEEE/CVF Conference on Computer Vision
  and Pattern Recognition (CVPR)}, pages 14455--14465, 2024.

\bibitem[Chen et~al.(2026{\natexlab{a}})Chen, Uy, Song, Ladhak, Murali, Qu,
  Birchfield, Blukis, and Tremblay]{chen2026spacetools}
Siyi Chen, Mikaela~Angelina Uy, Chan~Hee Song, Faisal Ladhak, Adithyavairavan
  Murali, Qing Qu, Stan Birchfield, Valts Blukis, and Jonathan Tremblay.
\newblock {SpaceTools}: Tool-augmented spatial reasoning via double interactive
  {RL}.
\newblock In \emph{Proceedings of the IEEE/CVF Conference on Computer Vision
  and Pattern Recognition (CVPR)}, pages 37109--37120, 2026{\natexlab{a}}.

\bibitem[Chen et~al.(2026{\natexlab{b}})Chen, Lu, Zheng, Li, He, Zhou, Shao,
  Zhuang, and Sheng]{chen2026gca}
Zeren Chen, Xiaoya Lu, Zhijie Zheng, Pengrui Li, Lehan He, Yijin Zhou, Jing
  Shao, Bohan Zhuang, and Lu Sheng.
\newblock Geometrically-constrained agent for spatial reasoning.
\newblock In \emph{Proceedings of the IEEE/CVF Conference on Computer Vision
  and Pattern Recognition (CVPR)}, pages 38689--38699, 2026{\natexlab{b}}.

\bibitem[Cheng et~al.(2024)Cheng, Yin, Fu, Guo, Yang, Kautz, Wang, and
  Liu]{cheng2024spatialrgpt}
An-Chieh Cheng, Hongxu Yin, Yang Fu, Qiushan Guo, Ruihan Yang, Jan Kautz,
  Xiaolong Wang, and Sifei Liu.
\newblock {SpatialRGPT}: Grounded spatial reasoning in vision-language models.
\newblock In \emph{Advances in Neural Information Processing Systems
  (NeurIPS)}, 2024.

\bibitem[Comanici et~al.(2025)Comanici, Bieber, Schaekermann, Pasupat,
  Sachdeva, Dhillon, Blistein, Ram, Zhang, Rosen, et~al.]{comanici2025gemini}
Gheorghe Comanici, Eric Bieber, Mike Schaekermann, Ice Pasupat, Noveen
  Sachdeva, Inderjit Dhillon, Marcel Blistein, Ori Ram, Dan Zhang, Evan Rosen,
  et~al.
\newblock Gemini 2.5: Pushing the frontier with advanced reasoning,
  multimodality, long context, and next generation agentic capabilities.
\newblock \emph{arXiv preprint arXiv:2507.06261}, 2025.

\bibitem[Dai et~al.(2017)Dai, Chang, Savva, Halber, Funkhouser, and
  Nie{\ss}ner]{dai2017scannet}
Angela Dai, Angel~X. Chang, Manolis Savva, Maciej Halber, Thomas Funkhouser,
  and Matthias Nie{\ss}ner.
\newblock {ScanNet}: Richly-annotated {3D} reconstructions of indoor scenes.
\newblock In \emph{Proceedings of the IEEE Conference on Computer Vision and
  Pattern Recognition (CVPR)}, 2017.

\bibitem[Deng et~al.(2025)Deng, Cao, Chen, and Hooi]{deng2025words}
Ailin Deng, Tri Cao, Zhirui Chen, and Bryan Hooi.
\newblock Words or vision: Do vision-language models have blind faith in text?
\newblock In \emph{Proceedings of the IEEE/CVF Conference on Computer Vision
  and Pattern Recognition (CVPR)}, pages 3867--3876, 2025.

\bibitem[Du et~al.(2026)Du, Ropero, Turkoz, Zhang, and Liu]{du2026graft}
Junqing Du, Fernando Ropero, Erkin Turkoz, Yanfeng Zhang, and Lu Liu.
\newblock {GraFT}: A training-free framework for spatial reasoning in
  multimodal large language models via {3D} scene graphs.
\newblock \emph{arXiv preprint arXiv:2609.03892}, 2026.

\bibitem[Efron and Tibshirani(1993)]{efron1993bootstrap}
Bradley Efron and Robert~J. Tibshirani.
\newblock \emph{An Introduction to the Bootstrap}.
\newblock Chapman \& Hall, 1993.

\bibitem[Fan et~al.(2026)Fan, Zhang, Li, Zhang, Chen, Hu, Wang, Wang, Qu, Zhou,
  Wang, Yan, Xu, Theiss, Chen, Li, Tu, Wang, and Ranjan]{fan2026vlm3r}
Zhiwen Fan, Jian Zhang, Renjie Li, Junge Zhang, Runjin Chen, Hezhen Hu, Kevin
  Wang, Peihao Wang, Huaizhi Qu, Shijie Zhou, Dilin Wang, Zhicheng Yan, Hongyu
  Xu, Justin Theiss, Tianlong Chen, Jiachen Li, Zhengzhong Tu, Zhangyang Wang,
  and Rakesh Ranjan.
\newblock {VLM-3R}: Vision-language models augmented with instruction-aligned
  {3D} reconstruction.
\newblock In \emph{Proceedings of the IEEE/CVF Conference on Computer Vision
  and Pattern Recognition (CVPR)}, pages 31054--31065, 2026.

\bibitem[Fischler and Bolles(1981)]{fischler1981ransac}
Martin~A. Fischler and Robert~C. Bolles.
\newblock Random sample consensus: A paradigm for model fitting with
  applications to image analysis and automated cartography.
\newblock \emph{Communications of the ACM}, 24\penalty0 (6):\penalty0 381--395,
  1981.

\bibitem[Gholami et~al.(2026)Gholami, Rezaei, Weimin, Mao, Zhou, Zhang, and
  Akbari]{gholami2026ego3d}
Mohsen Gholami, Ahmad Rezaei, Zhou Weimin, Sitong Mao, Shunbo Zhou, Yong Zhang,
  and Mohammad Akbari.
\newblock Spatial reasoning with vision-language models in ego-centric
  multi-view scenes.
\newblock In \emph{International Conference on Learning Representations
  (ICLR)}, 2026.

\bibitem[Gong et~al.(2026)Gong, Li, Tosi, Han, Mattoccia, Cai, and
  Poggi]{gong2026ov3r}
Ziren Gong, Xiaohan Li, Fabio Tosi, Jiawei Han, Stefano Mattoccia, Jianfei Cai,
  and Matteo Poggi.
\newblock {Ov3R}: Open-vocabulary semantic {3D} reconstruction from {RGB}
  videos.
\newblock In \emph{Proceedings of the IEEE/CVF Conference on Computer Vision
  and Pattern Recognition (CVPR)}, pages 34206--34216, 2026.

\bibitem[Gu et~al.(2024)Gu, Kuwajerwala, Morin, Jatavallabhula, Sen, Agarwal,
  Rivera, Paul, Ellis, Chellappa, Gan, de~Melo, Tenenbaum, Torralba, Shkurti,
  and Paull]{gu2024conceptgraphs}
Qiao Gu, Ali Kuwajerwala, Sacha Morin, Krishna~Murthy Jatavallabhula, Bipasha
  Sen, Aditya Agarwal, Corban Rivera, William Paul, Kirsty Ellis, Rama
  Chellappa, Chuang Gan, Celso~Miguel de Melo, Joshua~B. Tenenbaum, Antonio
  Torralba, Florian Shkurti, and Liam Paull.
\newblock {ConceptGraphs}: Open-vocabulary {3D} scene graphs for perception and
  planning.
\newblock In \emph{IEEE International Conference on Robotics and Automation
  (ICRA)}, pages 5021--5028, 2024.

\bibitem[Guo et~al.(2026)Guo, Hou, Ma, Tang, and Yang]{guo2026mssr}
Yejie Guo, Yunzhong Hou, Wufei Ma, Meng Tang, and Ming-Hsuan Yang.
\newblock Pursuing minimal sufficiency in spatial reasoning.
\newblock In \emph{International Conference on Learning Representations
  (ICLR)}, 2026.

\bibitem[Gupta and Kembhavi(2023)]{gupta2023visprog}
Tanmay Gupta and Aniruddha Kembhavi.
\newblock Visual programming: Compositional visual reasoning without training.
\newblock In \emph{Proceedings of the IEEE/CVF Conference on Computer Vision
  and Pattern Recognition (CVPR)}, pages 14953--14962, 2023.

\bibitem[Hoiem et~al.(2012)Hoiem, Chodpathumwan, and Dai]{hoiem2012diagnosing}
Derek Hoiem, Yodsawalai Chodpathumwan, and Qieyun Dai.
\newblock Diagnosing error in object detectors.
\newblock In \emph{Computer Vision -- {ECCV} 2012}, pages 340--353. Springer,
  2012.

\bibitem[Holm(1979)]{holm1979simple}
Sture Holm.
\newblock A simple sequentially rejective multiple test procedure.
\newblock \emph{Scandinavian Journal of Statistics}, 6\penalty0 (2):\penalty0
  65--70, 1979.

\bibitem[Hsu et~al.(2025)Hsu, Zhang, Wang, Tu, Li, Liu, Chen, Sun, and
  Kuo]{hsu2025openm3d}
Peng-Hao Hsu, Ke Zhang, Fu-En Wang, Tao Tu, Ming-Feng Li, Yu-Lun Liu, Albert
  Y.~C. Chen, Min Sun, and Cheng-Hao Kuo.
\newblock {OpenM3D}: Open vocabulary multi-view indoor {3D} object detection
  without human annotations.
\newblock In \emph{Proceedings of the IEEE/CVF International Conference on
  Computer Vision (ICCV)}, pages 8688--8698, 2025.

\bibitem[Hua et~al.(2026)Hua, Yin, Wu, Wang, Huang, and Liu]{hua2026trace}
Jiacheng Hua, Yishu Yin, Yuhang Wu, Tai Wang, Yifei Huang, and Miao Liu.
\newblock Unleashing spatial reasoning in multimodal large language models via
  textual representation guided reasoning.
\newblock In \emph{Proceedings of the Annual Meeting of the Association for
  Computational Linguistics (ACL)}, pages 13616--13637, 2026.

\bibitem[Huang et~al.(2026{\natexlab{a}})Huang, Zhang, Huang, Yang, and
  Tang]{huang2026consispace}
Ting Huang, Zhenyu Zhang, Wenyuan Huang, Jian Yang, and Hao Tang.
\newblock {ConsiSpace}: Learning geometric consistency matters for video
  spatial reasoning.
\newblock In \emph{European Conference on Computer Vision (ECCV)},
  2026{\natexlab{a}}.

\bibitem[Huang et~al.(2026{\natexlab{b}})Huang, Zhang, Li, Jia, Duan, Cheng,
  Cho, Wallingford, Soraki, Kim, Liu, Clay, Anderson, Han, Farhadi, Hariharan,
  Ren, and Krishna]{huang2026wilddet3d}
Weikai Huang, Jieyu Zhang, Sijun Li, Taoyang Jia, Jiafei Duan, Yunqian Cheng,
  Jaemin Cho, Matthew Wallingford, Rustin Soraki, Chris~Dongjoo Kim, Shuo Liu,
  Donovan Clay, Taira Anderson, Winson Han, Ali Farhadi, Bharath Hariharan,
  Zhongzheng Ren, and Ranjay Krishna.
\newblock {WildDet3D}: Scaling promptable {3D} detection in the wild.
\newblock \emph{arXiv preprint arXiv:2604.08626}, 2026{\natexlab{b}}.

\bibitem[Kuhn(1955)]{kuhn1955hungarian}
Harold~W. Kuhn.
\newblock The {H}ungarian method for the assignment problem.
\newblock \emph{Naval Research Logistics Quarterly}, 2\penalty0
  (1--2):\penalty0 83--97, 1955.

\bibitem[Li et~al.(2026{\natexlab{a}})Li, Li, Wang, Yan, Zhang, Chen, Hou,
  Jiang, Zhang, Shen, Lu, and Zhuang]{li2026viewspatial}
Dingming Li, Hongxing Li, Zixuan Wang, Yuchen Yan, Hang Zhang, Siqi Chen,
  Guiyang Hou, Shengpei Jiang, Wenqi Zhang, Yongliang Shen, Weiming Lu, and
  Yueting Zhuang.
\newblock {ViewSpatial-Bench}: Evaluating multi-perspective spatial
  localization in vision-language models.
\newblock In \emph{European Conference on Computer Vision (ECCV)},
  2026{\natexlab{a}}.

\bibitem[Li et~al.(2026{\natexlab{b}})Li, Cao, Tang, Xiang, Guo, Han, Xu, and
  Liang]{li2026geothinker}
Haoyuan Li, Qihang Cao, Tao Tang, Kun Xiang, Zihan Guo, Jianhua Han, Hang Xu,
  and Xiaodan Liang.
\newblock Thinking with geometry: Active geometry integration for spatial
  reasoning.
\newblock In \emph{Proceedings of the International Conference on Machine
  Learning (ICML)}, 2026{\natexlab{b}}.

\bibitem[Li et~al.(2026{\natexlab{c}})Li, Li, Wang, Yan, Wu, Zhang, Shen, Lu,
  Xiao, and Zhuang]{li2026spatialladder}
Hongxing Li, Dingming Li, Zixuan Wang, Yuchen Yan, Hang Wu, Wenqi Zhang,
  Yongliang Shen, Weiming Lu, Jun Xiao, and Yueting Zhuang.
\newblock {SpatialLadder}: Progressive training for spatial reasoning in
  vision-language models.
\newblock In \emph{International Conference on Learning Representations
  (ICLR)}, 2026{\natexlab{c}}.

\bibitem[Li et~al.(2025)Li, Song, Li, Yao, Guo, Xu, Liu, and
  Xiong]{li2025seetrek}
Pengteng Li, Pinhao Song, Wuyang Li, Huizai Yao, Weiyu Guo, Yijie Xu, Dugang
  Liu, and Hui Xiong.
\newblock {See\&Trek}: Training-free spatial prompting for multimodal large
  language model.
\newblock In \emph{Advances in Neural Information Processing Systems
  (NeurIPS)}, 2025.

\bibitem[Li et~al.(2026{\natexlab{d}})Li, Ma, Li, Li, Rong, Xu, Zhang, Zhao,
  and Huang]{li2026starr1}
Zongzhao Li, Zongyang Ma, Mingze Li, Songyou Li, Yu Rong, Tingyang Xu, Ziqi
  Zhang, Deli Zhao, and Wenbing Huang.
\newblock {STAR-R1}: Multi-view spatial transformation reasoning by reinforcing
  multimodal {LLMs}.
\newblock In \emph{Proceedings of the IEEE/CVF Conference on Computer Vision
  and Pattern Recognition (CVPR)}, pages 12041--12051, 2026{\natexlab{d}}.

\bibitem[Liu et~al.(2025)Liu, Dong, Wang, Ma, Tang, Tang, Rao, Ma, and
  Krishna]{liu2025coarse}
Benlin Liu, Yuhao Dong, Yiqin Wang, Zixian Ma, Yansong Tang, Luming Tang,
  Yongming Rao, Wei-Chiu Ma, and Ranjay Krishna.
\newblock Coarse correspondences boost spatial-temporal reasoning in multimodal
  language model.
\newblock In \emph{Proceedings of the Computer Vision and Pattern Recognition
  Conference (CVPR)}, pages 3783--3792, 2025.

\bibitem[Liu et~al.(2026{\natexlab{a}})Liu, Wu, Chi, Cai, Hung, Yu, Li, Hu,
  Rao, and Duan]{liu2026spatialttt}
Fangfu Liu, Diankun Wu, Jiawei Chi, Yimo Cai, Yi-Hsin Hung, Xumin Yu, Hao Li,
  Han Hu, Yongming Rao, and Yueqi Duan.
\newblock {Spatial-TTT}: Streaming visual-based spatial intelligence with
  test-time training.
\newblock In \emph{European Conference on Computer Vision (ECCV)},
  2026{\natexlab{a}}.

\bibitem[Liu et~al.(2024)Liu, Zeng, Ren, Li, Zhang, Yang, Jiang, Li, Yang, Su,
  Zhu, and Zhang]{liu2024groundingdino}
Shilong Liu, Zhaoyang Zeng, Tianhe Ren, Feng Li, Hao Zhang, Jie Yang, Qing
  Jiang, Chunyuan Li, Jianwei Yang, Hang Su, Jun Zhu, and Lei Zhang.
\newblock Grounding {DINO}: Marrying {DINO} with grounded pre-training for
  open-set object detection.
\newblock In \emph{Computer Vision -- {ECCV} 2024}, pages 38--55. Springer,
  2024.

\bibitem[Liu et~al.(2026{\natexlab{b}})Liu, Zhan, Zhou, Du, Liang, and
  Pfister]{liu2026sandboxvlm}
Yifan Liu, Fangneng Zhan, Kaichen Zhou, Yilun Du, Paul~Pu Liang, and Hanspeter
  Pfister.
\newblock Abstract {3D} perception for spatial intelligence in vision-language
  models.
\newblock In \emph{Proceedings of the IEEE/CVF Conference on Computer Vision
  and Pattern Recognition (CVPR)}, 2026{\natexlab{b}}.

\bibitem[Liu et~al.(2021)Liu, Lin, Cao, Hu, Wei, Zhang, Lin, and
  Guo]{liu2021swin}
Ze Liu, Yutong Lin, Yue Cao, Han Hu, Yixuan Wei, Zheng Zhang, Stephen Lin, and
  Baining Guo.
\newblock Swin transformer: Hierarchical vision transformer using shifted
  windows.
\newblock In \emph{IEEE/CVF International Conference on Computer Vision
  (ICCV)}, pages 10012--10022, 2021.

\bibitem[Liu et~al.(2026{\natexlab{c}})Liu, Du, Fu, Su, Ho, and
  Wang]{liu2026vlm2}
Zuntao Liu, Yi Du, Taimeng Fu, Shaoshu Su, Cherie Ho, and Chen Wang.
\newblock Vision-language memory for spatial reasoning.
\newblock In \emph{European Conference on Computer Vision (ECCV)},
  2026{\natexlab{c}}.

\bibitem[Luo et~al.(2026)Luo, Zhang, Yong, Dai, Wang, Ran, Shi, Sycara, and
  Xie]{luo2026pyspatial}
Zhanpeng Luo, Ce Zhang, Silong Yong, Cunxi Dai, Qianwei Wang, Haoxi Ran, Guanya
  Shi, Katia~P. Sycara, and Yaqi Xie.
\newblock {pySpatial}: Generating {3D} visual programs for zero-shot spatial
  reasoning.
\newblock In \emph{International Conference on Learning Representations
  (ICLR)}, 2026.

\bibitem[Ma et~al.(2024)Ma, Lu, Cheng, Trigoni, and Markham]{ma2024spatialpin}
Chenyang Ma, Kai Lu, Ta-Ying Cheng, Niki Trigoni, and Andrew Markham.
\newblock {SpatialPIN}: Enhancing spatial reasoning capabilities of
  vision-language models through prompting and interacting {3D} priors.
\newblock In \emph{Advances in Neural Information Processing Systems
  (NeurIPS)}, 2024.

\bibitem[Ma et~al.(2023)Ma, Yong, Zheng, Li, Liang, Zhu, and
  Huang]{ma2023sqa3d}
Xiaojian Ma, Silong Yong, Zilong Zheng, Qing Li, Yitao Liang, Song-Chun Zhu,
  and Siyuan Huang.
\newblock {SQA3D}: Situated question answering in {3D} scenes.
\newblock In \emph{International Conference on Learning Representations
  (ICLR)}, 2023.

\bibitem[Ma et~al.(2026)Ma, Sun, Chen, Bhalgat, Gu, Chang, Armeni, Laina, Peng,
  and Prisacariu]{ma2026do3dllm}
Xianzheng Ma, Tao Sun, Shuai Chen, Yash~Sanjay Bhalgat, Jindong Gu, Angel~X.
  Chang, Iro Armeni, Iro Laina, Songyou Peng, and Victor~Adrian Prisacariu.
\newblock Do {3D} large language models really understand {3D} spatial
  relationships?
\newblock In \emph{International Conference on Learning Representations
  (ICLR)}, 2026.

\bibitem[Majumdar et~al.(2024)Majumdar, Ajay, Zhang, Putta, Yenamandra, Henaff,
  Silwal, Mcvay, Maksymets, Arnaud, Yadav, Li, Newman, Sharma, Berges, Zhang,
  Agrawal, Bisk, Batra, Kalakrishnan, Meier, Paxton, Sax, and
  Rajeswaran]{majumdar2024openeqa}
Arjun Majumdar, Anurag Ajay, Xiaohan Zhang, Pranav Putta, Sriram Yenamandra,
  Mikael Henaff, Sneha Silwal, Paul Mcvay, Oleksandr Maksymets, Sergio Arnaud,
  Karmesh Yadav, Qiyang Li, Ben Newman, Mohit Sharma, Vincent Berges, Shiqi
  Zhang, Pulkit Agrawal, Yonatan Bisk, Dhruv Batra, Mrinal Kalakrishnan,
  Franziska Meier, Chris Paxton, Alexander Sax, and Aravind Rajeswaran.
\newblock {OpenEQA}: Embodied question answering in the era of foundation
  models.
\newblock In \emph{Proceedings of the IEEE/CVF Conference on Computer Vision
  and Pattern Recognition (CVPR)}, pages 16488--16498, 2024.

\bibitem[Mao et~al.(2025)Mao, Zhong, Fang, Zheng, Tang, Zhu, Tan, and
  Zhou]{mao2025spatiallm}
Yongsen Mao, Junhao Zhong, Chuan Fang, Jia Zheng, Rui Tang, Hao Zhu, Ping Tan,
  and Zihan Zhou.
\newblock {SpatialLM}: Training large language models for structured indoor
  modeling.
\newblock In \emph{Advances in Neural Information Processing Systems
  (NeurIPS)}, 2025.

\bibitem[Marsili et~al.(2025)Marsili, Agrawal, Yue, and
  Gkioxari]{marsili2025vadar}
Damiano Marsili, Rohun Agrawal, Yisong Yue, and Georgia Gkioxari.
\newblock Visual agentic {AI} for spatial reasoning with a dynamic {API}.
\newblock In \emph{Proceedings of the Computer Vision and Pattern Recognition
  Conference (CVPR)}, pages 19446--19455, 2025.

\bibitem[Miller(1995)]{miller1995wordnet}
George~A. Miller.
\newblock {WordNet}: A lexical database for {E}nglish.
\newblock \emph{Communications of the ACM}, 38\penalty0 (11):\penalty0 39--41,
  1995.

\bibitem[Nguyen et~al.(2024)Nguyen, Ngo, Kalogerakis, Gan, Tran, Pham, and
  Nguyen]{nguyen2024open3dis}
Phuc Nguyen, Tuan~Duc Ngo, Evangelos Kalogerakis, Chuang Gan, Anh Tran, Cuong
  Pham, and Khoi Nguyen.
\newblock {Open3DIS}: Open-vocabulary {3D} instance segmentation with {2D} mask
  guidance.
\newblock In \emph{Proceedings of the IEEE/CVF Conference on Computer Vision
  and Pattern Recognition (CVPR)}, pages 4018--4028, 2024.

\bibitem[{OpenAI}(2026)]{openai2026gpt6astra}
{OpenAI}.
\newblock {GPT-6 Astra} system card.
\newblock \url{https://deploymentsafety.openai.com/gpt-6-astra}, 2026.

\bibitem[Philion et~al.(2020)Philion, Kar, and Fidler]{philion2020pkl}
Jonah Philion, Amlan Kar, and Sanja Fidler.
\newblock Learning to evaluate perception models using planner-centric metrics.
\newblock In \emph{IEEE/CVF Conference on Computer Vision and Pattern
  Recognition (CVPR)}, pages 14055--14064, 2020.

\bibitem[Pichai et~al.(2025)Pichai, Hassabis, and
  Kavukcuoglu]{google2025gemini3}
Sundar Pichai, Demis Hassabis, and Koray Kavukcuoglu.
\newblock A new era of intelligence with {Gemini 3}.
\newblock
  \url{https://blog.google/products-and-platforms/products/gemini/gemini-3/},
  2025.

\bibitem[Qi et~al.(2026)Qi, Zhang, Fang, Wang, and Zhao]{qi2026gpt4scene}
Zhangyang Qi, Zhixiong Zhang, Ye Fang, Jiaqi Wang, and Hengshuang Zhao.
\newblock {GPT4Scene}: Understand {3D} scenes from videos with vision-language
  models.
\newblock In \emph{International Conference on Learning Representations
  (ICLR)}, 2026.

\bibitem[Rana et~al.(2023)Rana, Haviland, Garg, Abou-Chakra, Reid, and
  Suenderhauf]{rana2023sayplan}
Krishan Rana, Jesse Haviland, Sourav Garg, Jad Abou-Chakra, Ian Reid, and Niko
  Suenderhauf.
\newblock {SayPlan}: Grounding large language models using {3D} scene graphs
  for scalable robot task planning.
\newblock In \emph{Proceedings of The 7th Conference on Robot Learning (CoRL)},
  pages 23--72. PMLR, 2023.

\bibitem[Rozenberszki et~al.(2022)Rozenberszki, Litany, and
  Dai]{rozenberszki2022scannet200}
David Rozenberszki, Or Litany, and Angela Dai.
\newblock Language-grounded indoor {3D} semantic segmentation in the wild.
\newblock In \emph{European Conference on Computer Vision (ECCV)}, 2022.

\bibitem[Shi et~al.(2023)Shi, Chen, Misra, Scales, Dohan, Chi, Sch{\"a}rli, and
  Zhou]{shi2023distracted}
Freda Shi, Xinyun Chen, Kanishka Misra, Nathan Scales, David Dohan, Ed~H. Chi,
  Nathanael Sch{\"a}rli, and Denny Zhou.
\newblock Large language models can be easily distracted by irrelevant context.
\newblock In \emph{Proceedings of the 40th International Conference on Machine
  Learning (ICML)}, pages 31210--31227. PMLR, 2023.

\bibitem[Singh et~al.(2025)]{singh2025gpt5}
Aaditya Singh et~al.
\newblock {OpenAI} {GPT-5} system card.
\newblock \emph{arXiv preprint arXiv:2601.03267}, 2025.

\bibitem[Sur{\'\i}s et~al.(2023)Sur{\'\i}s, Menon, and
  Vondrick]{suris2023vipergpt}
D{\'\i}dac Sur{\'\i}s, Sachit Menon, and Carl Vondrick.
\newblock {ViperGPT}: Visual inference via python execution for reasoning.
\newblock In \emph{Proceedings of the IEEE/CVF International Conference on
  Computer Vision (ICCV)}, pages 11888--11898, 2023.

\bibitem[Takmaz et~al.(2023)Takmaz, Fedele, Sumner, Pollefeys, Tombari, and
  Engelmann]{takmaz2023openmask3d}
Ay{\c{c}}a Takmaz, Elisabetta Fedele, Robert~W. Sumner, Marc Pollefeys,
  Federico Tombari, and Francis Engelmann.
\newblock {OpenMask3D}: Open-vocabulary {3D} instance segmentation.
\newblock In \emph{Advances in Neural Information Processing Systems
  (NeurIPS)}, 2023.

\bibitem[Wang et~al.(2025)Wang, Chen, Karaev, Vedaldi, Rupprecht, and
  Novotny]{wang2025vggt}
Jianyuan Wang, Minghao Chen, Nikita Karaev, Andrea Vedaldi, Christian
  Rupprecht, and David Novotny.
\newblock {VGGT}: Visual geometry grounded transformer.
\newblock In \emph{Proceedings of the Computer Vision and Pattern Recognition
  Conference (CVPR)}, pages 5294--5306, 2025.

\bibitem[Wang et~al.(2026{\natexlab{a}})Wang, Yin, Zhang, Zhang, Wang, Wang,
  Zhang, Chandrasegaran, Liu, Krishna, Xie, Wu, Fei-Fei, and
  Li]{wang2026mindcube}
Qineng Wang, Baiqiao Yin, Pingyue Zhang, Jianshu Zhang, Kangrui Wang, Zihan
  Wang, Jieyu Zhang, Keshigeyan Chandrasegaran, Han Liu, Ranjay Krishna,
  Saining Xie, Jiajun Wu, Li Fei-Fei, and Manling Li.
\newblock Spatial mental modeling from limited views.
\newblock In \emph{International Conference on Learning Representations
  (ICLR)}, 2026{\natexlab{a}}.

\bibitem[Wang et~al.(2026{\natexlab{b}})Wang, Zhou, Zhu, Chang, Zhou, Li, Chen,
  Pang, Shen, and He]{wang2026pi3}
Yifan Wang, Jianjun Zhou, Haoyi Zhu, Wenzheng Chang, Yang Zhou, Zizun Li, Junyi
  Chen, Jiangmiao Pang, Chunhua Shen, and Tong He.
\newblock $\pi^3$: Permutation-equivariant visual geometry learning.
\newblock In \emph{International Conference on Learning Representations
  (ICLR)}, 2026{\natexlab{b}}.

\bibitem[Werby et~al.(2024)Werby, Huang, B{\"u}chner, Valada, and
  Burgard]{werby2024hovsg}
Abdelrhman Werby, Chenguang Huang, Martin B{\"u}chner, Abhinav Valada, and
  Wolfram Burgard.
\newblock Hierarchical open-vocabulary {3D} scene graphs for language-grounded
  robot navigation.
\newblock In \emph{Proceedings of Robotics: Science and Systems (RSS)}, 2024.

\bibitem[Wu et~al.(2025)Wu, Liu, Hung, and Duan]{wu2025spatialmllm}
Diankun Wu, Fangfu Liu, Yi-Hsin Hung, and Yueqi Duan.
\newblock {Spatial-MLLM}: Boosting {MLLM} capabilities in visual-based spatial
  intelligence.
\newblock In \emph{Advances in Neural Information Processing Systems
  (NeurIPS)}, 2025.

\bibitem[Xiao et~al.(2024)Xiao, Wu, Xu, Dai, Hu, Lu, Zeng, Liu, and
  Yuan]{xiao2024florence2}
Bin Xiao, Haiping Wu, Weijian Xu, Xiyang Dai, Houdong Hu, Yumao Lu, Michael
  Zeng, Ce Liu, and Lu Yuan.
\newblock {Florence-2}: Advancing a unified representation for a variety of
  vision tasks.
\newblock In \emph{Proceedings of the IEEE/CVF Conference on Computer Vision
  and Pattern Recognition (CVPR)}, pages 4818--4829, 2024.

\bibitem[Yan et~al.(2024)Yan, Zhang, Zhu, and Wang]{yan2024maskclustering}
Mi Yan, Jiazhao Zhang, Yan Zhu, and He Wang.
\newblock {MaskClustering}: View consensus based mask graph clustering for
  open-vocabulary {3D} instance segmentation.
\newblock In \emph{Proceedings of the IEEE/CVF Conference on Computer Vision
  and Pattern Recognition (CVPR)}, pages 28274--28284, 2024.

\bibitem[Yang et~al.(2025{\natexlab{a}})Yang, Yang, Gupta, Han, Fei-Fei, and
  Xie]{yang2025thinking}
Jihan Yang, Shusheng Yang, Anjali~W. Gupta, Rilyn Han, Li Fei-Fei, and Saining
  Xie.
\newblock Thinking in space: How multimodal large language models see,
  remember, and recall spaces.
\newblock In \emph{Proceedings of the Computer Vision and Pattern Recognition
  Conference (CVPR)}, pages 10632--10643, 2025{\natexlab{a}}.

\bibitem[Yang et~al.(2026{\natexlab{a}})Yang, Zhao, Pan, Yang, Zhang, Xu, Li,
  and Xie]{yang2026cambrianp}
Jihan Yang, Zifan Zhao, Xichen Pan, Shusheng Yang, Junyi Zhang, Hu Xu,
  Shang-Wen Li, and Saining Xie.
\newblock {Cambrian-P}: Pose-grounded video understanding.
\newblock In \emph{European Conference on Computer Vision (ECCV)},
  2026{\natexlab{a}}.

\bibitem[Yang et~al.(2026{\natexlab{b}})Yang, Xu, Xie, Yang, Li, Lin, Zhu,
  Chen, Duan, Yue, Lin, Wang, and Pang]{yang2026mmsi}
Sihan Yang, Runsen Xu, Yiman Xie, Sizhe Yang, Mo Li, Jingli Lin, Chenming Zhu,
  Xiaochen Chen, Haodong Duan, Xiangyu Yue, Dahua Lin, Tai Wang, and Jiangmiao
  Pang.
\newblock {MMSI-Bench}: A benchmark for multi-image spatial intelligence.
\newblock In \emph{International Conference on Learning Representations
  (ICLR)}, 2026{\natexlab{b}}.

\bibitem[Yang et~al.(2026{\natexlab{c}})Yang, Yang, Huang, Brown, Yang, Yu,
  Tong, Zheng, Xu, Wang, Fergus, LeCun, Fei-Fei, and Xie]{yang2026cambrians}
Shusheng Yang, Jihan Yang, Pinzhi Huang, Ellis Brown, Zihao Yang, Yue Yu,
  Shengbang Tong, Zihan Zheng, Yifan Xu, Muhan Wang, Rob Fergus, Yann LeCun, Li
  Fei-Fei, and Saining Xie.
\newblock {Cambrian-S}: Towards spatial supersensing in video.
\newblock In \emph{International Conference on Learning Representations
  (ICLR)}, 2026{\natexlab{c}}.

\bibitem[Yang et~al.(2025{\natexlab{b}})Yang, Piccinelli, Segu, Li, Huang, Fu,
  Pollefeys, Blum, and Bauer]{yang2025mood}
Yung-Hsu Yang, Luigi Piccinelli, Mattia Segu, Siyuan Li, Rui Huang, Yuqian Fu,
  Marc Pollefeys, Hermann Blum, and Zuria Bauer.
\newblock {3D-MOOD}: Lifting {2D} to {3D} for monocular open-set object
  detection.
\newblock In \emph{Proceedings of the IEEE/CVF International Conference on
  Computer Vision (ICCV)}, pages 7429--7439, 2025{\natexlab{b}}.

\bibitem[Yeshwanth et~al.(2023)Yeshwanth, Liu, Nie{\ss}ner, and
  Dai]{yeshwanth2023scannetpp}
Chandan Yeshwanth, Yueh-Cheng Liu, Matthias Nie{\ss}ner, and Angela Dai.
\newblock {ScanNet++}: A high-fidelity dataset of {3D} indoor scenes.
\newblock In \emph{Proceedings of the IEEE/CVF International Conference on
  Computer Vision (ICCV)}, 2023.

\bibitem[Yuan et~al.(2026)Yuan, Kumar, and Wang]{yuan2026gr3d}
Jiangye Yuan, Gowri Kumar, and Baoyuan Wang.
\newblock Boosting {MLLM} spatial reasoning with geometrically referenced {3D}
  scene representations.
\newblock \emph{arXiv preprint arXiv:2603.08592}, 2026.

\bibitem[Zhang et~al.(2025{\natexlab{a}})Zhang, Jiang, Yao, Sun, Zhang, Zhao,
  Li, Zhu, and Yang]{zhang2025detany3d}
Hanxue Zhang, Haoran Jiang, Qingsong Yao, Yanan Sun, Renrui Zhang, Hao Zhao,
  Hongyang Li, Hongzi Zhu, and Zetong Yang.
\newblock Detect anything {3D} in the wild.
\newblock In \emph{Proceedings of the IEEE/CVF International Conference on
  Computer Vision (ICCV)}, pages 5048--5059, 2025{\natexlab{a}}.

\bibitem[Zhang et~al.(2025{\natexlab{b}})Zhang, Chen, Zhou, Xu, Huang, Mei,
  Chen, Yuan, Cai, Huang, Quan, Xu, and Zhang]{zhang2025spar}
Jiahui Zhang, Yurui Chen, Yanpeng Zhou, Yueming Xu, Ze Huang, Jilin Mei, Junhui
  Chen, Yu-Jie Yuan, Xinyue Cai, Guowei Huang, Xingyue Quan, Hang Xu, and Li
  Zhang.
\newblock From flatland to space: Teaching vision-language models to perceive
  and reason in {3D}.
\newblock In \emph{Advances in Neural Information Processing Systems (NeurIPS),
  Datasets and Benchmarks Track}, 2025{\natexlab{b}}.

\bibitem[Zhu et~al.(2025)Zhu, Wang, Xie, Gu, Ding, Yang, and
  Jiang]{zhu2025struct2d}
Fangrui Zhu, Hanhui Wang, Yiming Xie, Jing Gu, Tianye Ding, Jianwei Yang, and
  Huaizu Jiang.
\newblock {Struct2D}: A perception-guided framework for spatial reasoning in
  {MLLMs}.
\newblock In \emph{Advances in Neural Information Processing Systems
  (NeurIPS)}, 2025.

\bibitem[Zust et~al.(2025)Zust, Cabon, Marrie, Antsfeld, Chidlovskii, Revaud,
  and Csurka]{zust2025panst3r}
Lojze Zust, Yohann Cabon, Juliette Marrie, Leonid Antsfeld, Boris Chidlovskii,
  Jerome Revaud, and Gabriela Csurka.
\newblock {PanSt3R}: Multi-view consistent panoptic segmentation.
\newblock In \emph{Proceedings of the IEEE/CVF International Conference on
  Computer Vision (ICCV)}, pages 5856--5866, 2025.

\end{thebibliography}
}

\clearpage
\maketitlesupplementary
\appendix
\renewcommand{\thetable}{S\arabic{table}}
\renewcommand{\thefigure}{S\arabic{figure}}
\setcounter{table}{0}\setcounter{figure}{0}
\renewcommand{\theHtable}{S\arabic{table}}\renewcommand{\theHfigure}{S\arabic{figure}}
\renewcommand{\topfraction}{0.9}\renewcommand{\dbltopfraction}{0.9}\renewcommand{\bottomfraction}{0.8}
\renewcommand{\textfraction}{0.07}\renewcommand{\floatpagefraction}{0.85}\renewcommand{\dblfloatpagefraction}{0.85}
\setcounter{topnumber}{4}\setcounter{dbltopnumber}{3}\setcounter{bottomnumber}{2}\setcounter{totalnumber}{6}

This supplement gives the details behind the paper: how we build the list (\cref{supp:list}), how we evaluate (\cref{supp:eval}), the paper's tables with 95\% intervals (\cref{supp:tables}), how we ran the other methods (\cref{supp:methods}), and the controlled error study with tests on real lists (\cref{supp:study}). Section, table and figure numbers without a letter or an S refer to the paper.

\section{Building the list}
\label{supp:list}

\Cref{tab:params} lists every setting of the pipeline.
\begin{table*}[t]
\centering
\footnotesize
\setlength{\tabcolsep}{4pt}
\renewcommand{\arraystretch}{1.1}
\begin{minipage}[t]{0.49\textwidth}
\centering
\begin{tabular}[t]{@{}ll@{}}
\toprule
Setting & Value \\
\midrule
\multicolumn{2}{@{}l}{\textit{Geometry (Pi3X~\cite{wang2026pi3})}} \\
Input frames & two sharpest per second, at most 106 \\
Pixel confidence & $>$0.6\smash{$^\dagger$}, not on a depth edge \\
Relative rule & also $\geq$ the median, if $<$1/2 of pixels pass \\
Later stages & pixel confidence $>$0.5 \\
\midrule
\multicolumn{2}{@{}l}{\textit{Names and segmentation}} \\
Fixed vocabulary & 157 categories \\
Caption names & 60 most frequent (Florence-2~\cite{xiao2024florence2}) \\
SAM~3~\cite{carion2026sam3} mask score & $\geq$0.5 \\
NMS across names & mask IoU $\geq$0.7 \\
\midrule
\multicolumn{2}{@{}l}{\textit{Boxes and floor}} \\
Box fit & gravity-aligned, 0.5\% of the extent cut per side \\
Floor snap & bottom $<$0.3\,m above the floor \\
Floor area & 5\,cm voxels, 10\,cm grid, largest region, holes filled \\
\bottomrule
\end{tabular}
\end{minipage}\hfill
\begin{minipage}[t]{0.49\textwidth}
\centering
\begin{tabular}[t]{@{}ll@{}}
\toprule
Setting & Value \\
\midrule
\multicolumn{2}{@{}l}{\textit{Lifting and association}} \\
Depth outliers & $>$3 MAD from the mask's median depth \\
Pixels per mask & $\geq$50 \\
Voxel size & 5\,cm \\
Link & $\geq$1/4 of the smaller voxel set, centres $\leq$1.5\,m apart \\
Same-name merge & box overlap $\geq$1/2 \\
Object kept & $\geq$100 points in $\geq$2 frames, or $\geq$300 points \\
\midrule
\multicolumn{2}{@{}l}{\textit{Matching to ground truth (analyses only, \cref{supp:eval})}} \\
Alignment & RANSAC~\cite{fischler1981ransac} on objects with matching names \\
Inliers & within 0.5\,m; $\geq$4 from $\geq$3 categories, or $\geq$8 from 2 \\
Scale & between 0.5 and 2 \\
Object found & centres $\leq$0.5\,m apart, or centre inside the box \\
\bottomrule
\end{tabular}
\end{minipage}
\caption{\textbf{Every setting of the pipeline} and of the matching to ground truth. \smash{$^\dagger$}0.4 on the ARKitScenes~\cite{baruch2021arkitscenes} videos of VSI-Bench~\cite{yang2025thinking}.}
\label{tab:params}
\end{table*}

The fixed vocabulary of Sec.~3.2 is:
{\scriptsize chair, armchair, office chair, stool, bench, sofa, ottoman, table, desk, coffee table, side table, nightstand, dresser, cabinet, cupboard, wardrobe, shelf, bookshelf, drawer, tv stand, bed, crib, mattress, headboard, television, monitor, computer, laptop, keyboard, computer mouse, computer tower, printer, speaker, telephone, tablet, projector, router, fan, heater, radiator, air conditioner, refrigerator, microwave, oven, stove, range hood, dishwasher, washing machine, dryer, kettle, toaster, coffee maker, blender, vacuum cleaner, lamp, floor lamp, table lamp, ceiling light, chandelier, light switch, power outlet, power strip, smoke detector, toilet, sink, bathtub, shower, mirror, towel, toilet paper, soap dispenser, bath mat, trash can, laundry basket, counter, kitchen island, cutting board, pot, pan, plate, bowl, cup, mug, bottle, jar, dish rack, paper towel, pillow, cushion, blanket, curtain, blinds, rug, clothes, jacket, hat, box, basket, bin, bag, backpack, suitcase, crate, container, shoe rack, coat rack, coat hanger, picture, painting, poster, clock, plant, potted plant, vase, candle, calendar, whiteboard, bulletin board, sign, door, window, book, magazine, paper, notebook, toy, ball, guitar, piano, umbrella, shoes, boots, helmet, bicycle, ladder, toolbox, fire extinguisher, tray, stand, mat, treadmill, exercise ball, easel, file cabinet, mailbox, cart, bucket, broom, iron, ironing board, sewing machine, globe, aquarium, pet bed, stroller, car seat, baby gate.}
For each scene, we add the 60 object names that occur most often in its Florence-2~\cite{xiao2024florence2} captions. A name is the main noun of an object phrase in a caption, or its last two words if they form a name of the vocabulary. Words for structure, such as wall, floor or room, are skipped.

When masks overlap, each pixel goes to the smallest mask, so that a large mask in the background does not take over the objects in front of it. Two masks of the same frame are merged only if they touch in 3D. For the merge rules of Sec.~3.2, two names count as the same if they have the same base form or the same main noun, such as office chair and chair; we use no synonym list. After the merging, small fragments inside an object join it, and so do overlapping objects with the same name. No step uses appearance features or the VLM.

Boxes that end just above the floor are extended down to it, because the bottom of an object is usually hidden.

We developed the pipeline on VSI-Bench, so we also measured the gain on the 220 scenes that played no part in any decision. The gain on these 220 scenes is 5.5 points, against 6.0 on all scenes.

\begin{figure*}[t]
\centering
\begin{minipage}{0.98\textwidth}
\scriptsize
\begin{verbatim}
### SYSTEM INSTRUCTION
You are an expert in 3D Spatial Intelligence.
[32 frames]
Answer each question by correlating the **video walkthrough** (the frames above, in temporal order) with the
**3D object list** below.

COORDINATE SYSTEM: metres, gravity-aligned. z is up and the floor is at z = 0; x and y are horizontal.
Each object has: label; center [x, y, z]; size [length, width, height] (length and width are horizontal, height is
vertical).
room_footprint_m2 = floor area covered by the reconstruction of the scene.
Distances between objects are measured between their closest points (box surfaces), not between centers.
The list is machine-generated: it can miss objects, contain extra ones, or split one object into several entries.
Verify with the frames.

### SCENE METADATA (JSON)
{"room_footprint_m2":18.92,"objects":[
{"label":"air conditioner","center":[2.17,-1.63,1.36],"size":[0.44,0.12,0.54]},
{"label":"armrest","center":[-1.17,0.64,0.31],"size":[0.72,0.21,0.61]},
...
{"label":"lamp","center":[1.46,1.97,1.15],"size":[0.43,0.16,0.65]},
...
{"label":"window","center":[2.2,-0.86,1.55],"size":[1.15,0.21,1.19]},
{"label":"window","center":[-1.75,-0.17,1.54],"size":[0.21,0.02,0.31]}]}

### TASKS
Answer each question below. For each question, output your answer in <answer>...</answer> tags.
Use exactly one <answer> block per question, in the same order as the questions.
For multiple-choice questions, answer with the option letter only.
For numeric questions, answer with a single number only (no units, no words).

### QUESTION 1
Measuring from the closest point of each object, what is the distance between the lamp and the window (in meters)?

### OUTPUT
CRITICAL: Provide exactly 1 <answer>...</answer> blocks in order: first <answer> for QUESTION 1,
second for QUESTION 2, etc. Do not reorder.
\end{verbatim}
\end{minipage}
\caption{\textbf{The prompt with our list} for the question of Fig.~1 (ScanNet~\cite{dai2017scannet} scene0616\_01), shortened. The list is sorted by label and sent as one line; only part of it is shown. The output instructions are generic; every request holds one question. With this prompt, Gemini~3.8~Flash~\cite{google2025gemini3} answers 2.11\,m; the ground truth is 2.2\,m.}
\label{fig:prompt}
\end{figure*}

Building a list takes about three minutes on one A100 GPU per VSI-Bench video, plus CPU time. Answers with our list are also cheaper (Sec.~4.2). Compared with the video alone, Gemini~3.8~Flash~\cite{google2025gemini3} thinks for 8.1k instead of 17.5k tokens per question. The list adds 7.7k input tokens, but an answer still costs about 40\% less, \$0.0145 instead of \$0.0254 at batch prices, over 1,585 VSI-Bench questions.

\section{Evaluation}
\label{supp:eval}

\Cref{tab:supp-reasoners} gives the settings of every VLM. The 800 VSI-Bench questions (100 per task) were drawn at random from 1,587 questions that we sampled beforehand. None of them comes from the ten scenes we used to write the prompt.

\begin{table*}[t]
\centering
\footnotesize
\setlength{\tabcolsep}{5pt}
\renewcommand{\arraystretch}{1.1}
\begin{tabular}{@{}lll@{}}
\toprule
VLM & Model & Settings \\
\midrule
Gemini~3.5~Flash-Lite~\cite{comanici2025gemini} & \texttt{gemini-3.5-flash-lite} & temperature 0, top-$p$ 1, thinking level high, default output limit \\
Gemini~3.1~Pro~\cite{comanici2025gemini} & \texttt{gemini-3.1-pro-preview} & temperature 0, top-$p$ 1, thinking level high, default output limit \\
Gemini~3.8~Flash~\cite{google2025gemini3} & \texttt{gemini-3.8-flash} & temperature 0, top-$p$ 1, thinking level high, default output limit \\
GPT-5~\cite{singh2025gpt5} & \texttt{gpt-5-2025-08-07} & reasoning effort medium, at most 32,000 output tokens \\
GPT-6 Astra~\cite{openai2026gpt6astra} & \texttt{gpt-6-astra} & reasoning effort medium, at most 32,000 output tokens \\
Qwen3-VL-8B-Thinking~\cite{bai2025qwen3vl} & released weights, bf16 & its own sampling (temperature 1.0, top-$p$ 0.95, top-$k$ 20), at most 8,192 new tokens \\
Qwen3-VL-32B-Instruct~\cite{bai2025qwen3vl} & released weights, bf16 & greedy decoding, at most 64 new tokens, no thinking \\
\bottomrule
\end{tabular}
\caption{\textbf{VLM settings.} Every VLM receives 32 frames at native resolution, evenly spaced over a video, or all images of an image set, and answers one question per request. Qwen3-VL-8B-Thinking draws one seeded sample; its answer is read after its reasoning, and a request without an answer counts as wrong.}
\label{tab:supp-reasoners}
\end{table*}

For the intervals of Sec.~4.1, the bootstrap~\cite{efron1993bootstrap} resamples scenes within each source dataset of VSI-Bench, 10,000 times (2,000 times for Tab.~1). Each condition was run once, and even at temperature 0 an answer can change between runs. The intervals therefore show the effect of which questions were sampled, not of this run-to-run change. Each interval is for its own result alone. Only the nine hypotheses of the error study were tested together, with Holm's method~\cite{holm1979simple}: it makes each test stricter, so that the chance of calling any of the nine supported by mistake stays below 5\%.

\Cref{tab:metrics_real,tab:oracle} compare a list with the ground-truth list of its scene. Both lists are gravity-aligned, so they differ only by a rotation about the vertical axis, a horizontal shift and a scale. We fit these with RANSAC~\cite{fischler1981ransac} on pairs of objects with matching names (thresholds in \cref{tab:params}). A name matches a ground-truth (GT) category if the two are equal after normalisation, if the category is the last word or words of the name, or if their most common WordNet~\cite{miller1995wordnet} senses agree. Within a category, we pair list and ground-truth objects with the Hungarian method~\cite{kuhn1955hungarian} on centre distance. The fitted scale also measures the scale error of our reconstruction. Its median is below 6\% on each source dataset of VSI-Bench: ScanNet, ARKitScenes~\cite{baruch2021arkitscenes} and ScanNet++~\cite{yeshwanth2023scannetpp}.

\section{The paper's tables with intervals}
\label{supp:tables}

\Cref{tab:t12,tab:t4,tab:t5,tab:t6} repeat Tables 1, 2, 4, 5 and 6 of the paper with 95\% intervals. In \cref{tab:t12,tab:t4}, bold marks the better result of each comparison.

\begin{table}[t]
\centering
\footnotesize
\setlength{\tabcolsep}{3pt}
\begin{tabular}{@{}lccc@{}}
\toprule
 & Video & + our list & Gain [95\% interval] \\
\midrule
\multicolumn{4}{@{}l}{\textit{Table 1: benchmarks}} \\
VSI-Bench~\cite{yang2025thinking} & 67.5 & \textbf{73.6} & 6.0 [4.2, 7.9] \\
VSI-Bench-Debiased~\cite{brown2026debiased} & 65.3 & \textbf{71.6} & 6.3 [4.0, 8.8] \\
MMSI-Bench~\cite{yang2026mmsi} & 64.1 & \textbf{72.2} & 8.1 [5.0, 11.4] \\
SPBench-MV~\cite{li2026spatialladder} & 72.8 & \textbf{78.0} & 5.2 [0.2, 9.8] \\
MindCube-Tiny~\cite{wang2026mindcube} & 92.4 & \textbf{94.3} & 1.9 [\ensuremath{-}0.1, 3.9] \\
\midrule
\multicolumn{4}{@{}l}{\textit{Table 2: VSI-Bench by task}} \\
Room size & 54.2 & \textbf{75.2} & 21.0 [16.7, 25.4] \\
Abs.\ distance & 52.2 & \textbf{69.7} & 17.5 [14.0, 21.0] \\
Route plan & 79.9 & \textbf{91.8} & 11.9 [6.2, 18.0] \\
Rel.\ direction & 76.4 & \textbf{84.4} & 8.0 [3.2, 12.9] \\
Appear.\ order & 73.0 & \textbf{74.3} & 1.3 [\ensuremath{-}2.7, 5.2] \\
Counting & \textbf{56.9} & 56.0 & \ensuremath{-}0.8 [\ensuremath{-}4.6, 2.8] \\
Rel.\ distance & \textbf{71.3} & 68.2 & \ensuremath{-}3.1 [\ensuremath{-}9.2, 3.0] \\
Object size & \textbf{76.5} & 69.1 & \ensuremath{-}7.4 [\ensuremath{-}9.7, \ensuremath{-}5.0] \\
\bottomrule
\end{tabular}
\caption{\textbf{Tables 1 and 2 of the paper with intervals.} Gemini~3.8~Flash with the same frames: accuracy (\%) and gain in points with its 95\% interval. Bold: the higher accuracy.}
\label{tab:t12}
\end{table}
\begin{table}[t]
\centering
\scriptsize
\setlength{\tabcolsep}{2pt}
\begin{tabular}{@{}lccc@{}}
\toprule
Method & \begin{tabular}[b]{@{}c@{}}Gain over\\video\end{tabular} & Our lead & \begin{tabular}[b]{@{}c@{}}Abs.\\dist.\end{tabular} \\
\midrule
Cognitive map~\cite{yang2025thinking} & 0.0 [\ensuremath{-}2.3, 2.4] & 5.9 [2.8, 9.0] & 6.7 \\
TRACE~\cite{hua2026trace} & \ensuremath{-}1.7 [\ensuremath{-}3.9, 0.5] & 7.6 [4.4, 10.8] & 1.9 \\
ConceptGraphs~\cite{gu2024conceptgraphs} & 0.9 [\ensuremath{-}1.5, 3.3] & 5.1 [2.0, 8.2] & 3.2 \\
PanSt3R~\cite{zust2025panst3r} & 1.9 [\ensuremath{-}0.7, 4.6] & 4.0 [1.0, 7.0] & 2.8 \\
3D-MOOD~\cite{yang2025mood} & 1.1 [\ensuremath{-}1.5, 3.7] & 4.8 [1.6, 8.0] & 9.3 \\
3D-MOOD, fused across frames & \ensuremath{-}0.1 [\ensuremath{-}2.7, 2.4] & 6.1 [2.9, 9.4] & 7.9 \\
WildDet3D~\cite{huang2026wilddet3d} & 2.5 [\ensuremath{-}0.0, 5.0] & 3.5 [0.5, 6.5] & 9.0 \\
SpatialLM~\cite{mao2025spatiallm} & 3.5 [0.8, 6.2] & 2.4 [\ensuremath{-}0.9, 5.7] & 9.8 \\
ConceptGraphs, our input & 5.0 [1.9, 8.0] & 0.9 [\ensuremath{-}1.4, 3.2] & 24.4 \\
\midrule
Our list & \textbf{5.9} [3.0, 9.0] & -- & \textbf{28.9} \\
Ground-truth list & 26.2 [23.3, 29.3] & \ensuremath{-}20.3 [\ensuremath{-}23.3, \ensuremath{-}17.4] & 49.7 \\
\bottomrule
\end{tabular}
\caption{\textbf{Table 4 of the paper with intervals.} Gemini~3.8~Flash on the 800 VSI-Bench questions, in points, with 95\% intervals. \emph{Our lead}: accuracy with our list minus accuracy with the method; it is negative where the method is ahead. \emph{Abs.\ dist.}: gain over the video on the 100 absolute-distance questions. Bold: best, not counting the ground-truth list.}
\label{tab:t4}
\end{table}
\begin{table}[t]
\centering
\scriptsize
\setlength{\tabcolsep}{3pt}
\begin{tabular}{@{}lc@{}}
\toprule
Change to our list & Accuracy change [95\% interval] \\
\midrule
Names only, without positions or sizes & \ensuremath{-}8.6 [\ensuremath{-}11.7, \ensuremath{-}5.4] \\
Only names that map to VSI-Bench categories & \ensuremath{-}2.6 [\ensuremath{-}4.6, \ensuremath{-}0.5] \\
Half of the objects removed at random & \ensuremath{-}3.8 [\ensuremath{-}6.4, \ensuremath{-}1.2] \\
Fixed vocabulary only, no caption names & \ensuremath{-}0.5 [\ensuremath{-}2.2, 1.3] \\
Stricter detection threshold & \ensuremath{-}0.5 [\ensuremath{-}2.3, 1.4] \\
Only objects seen in at least three frames & \ensuremath{-}0.2 [\ensuremath{-}1.9, 1.5] \\
List only, without the frames (text only) & \ensuremath{-}11.8 [\ensuremath{-}14.8, \ensuremath{-}8.9] \\
No list (video alone) & \ensuremath{-}5.9 [\ensuremath{-}9.0, \ensuremath{-}2.9] \\
\bottomrule
\end{tabular}
\caption{\textbf{Table 5 of the paper with intervals.} Gemini~3.8~Flash on the 800 questions: change in accuracy against the full list (74.5\%), in points.}
\label{tab:t5}
\end{table}

Some methods give no list for some scenes. There the VLM answers from the video alone (Sec.~4.3). A method that ran but found no objects gives an empty list instead.

In \cref{tab:t5}, \emph{names only} also drops the floor area, and its prompt says that the list has no positions or sizes. Names are mapped to VSI-Bench categories with the rule of \cref{supp:eval}. The \emph{stricter detection threshold} drops the quarter of the masks with the lowest SAM~3 scores.

\begin{table}[t]
\centering
\footnotesize
\setlength{\tabcolsep}{4pt}
\begin{tabular}{@{}lc@{}}
\toprule
Change to the GT list & Lost by the VLM [95\% interval] \\
\midrule
Missing & 17.7 [15.0, 20.6] \\
Duplicated & 6.3 [4.5, 8.0] \\
Size $\times$2 & 23.9 [22.2, 25.6] \\
Shifted 1\,m & 14.5 [12.2, 16.9] \\
Shifted 0.5\,m & 7.6 [5.7, 9.5] \\
Unrelated objects added & 1.7 [0.3, 3.1] \\
\midrule
No list & 26.2 [22.7, 29.7] \\
\bottomrule
\end{tabular}
\caption{\textbf{Table 6 of the paper with intervals.} Accuracy lost by Gemini~3.8~Flash against the intact ground-truth list, in points over the six tasks, on the 592 questions that ask about specific objects (572 for unrelated objects). The detection columns of Tab.~6 use the lists of the same questions.}
\label{tab:t6}
\end{table}
\begin{table*}[t]
\centering
\footnotesize
\setlength{\tabcolsep}{4pt}
\renewcommand{\arraystretch}{1.1}
\begin{tabular}{@{}lll@{}}
\toprule
Error & Change to the ground-truth list & Levels \\
\midrule
\multicolumn{3}{@{}l}{\textit{The asked-about object}} \\
Missing & removed; for route planning, appearance order and answer options, every object of its category & -- \\
Duplicated & a copy 0.62\,m away in a random horizontal direction (the median offset of duplicates in our lists) & -- \\
Shifted & centre moved horizontally in a random direction & 0.1, 0.25, 0.5, 1\,m \\
Size & all three sides scaled about the centre & $\times$0.5, 0.8, 1.25, 2 \\
\midrule
\multicolumn{3}{@{}l}{\textit{The rest of the scene}} \\
Unrelated objects & categories in neither the scene nor the question, at their median size, at random free spots on the floor & +50\% \\
Half of the other objects & objects the question does not name, removed at random & 50\% \\
Global scale & the whole list scaled about its centre (10th and 90th percentiles of our lists' scale error) & $\times$0.925, 1.078 \\
Random removal & objects removed at random, whatever the question asks & 20\%, 50\% \\
Another room & the ground-truth list of another scene of the same dataset & -- \\
\bottomrule
\end{tabular}
\caption{\textbf{The errors of the study.} Each request applies one error at one level to the intact ground-truth list of the question's scene. Table~6 uses a missing, duplicated, doubled or shifted (0.5 and 1\,m) asked-about object and unrelated objects; the larger study uses every row and level.}
\label{tab:operators}
\end{table*}
\begin{table*}[t]
\centering
\scriptsize
\setlength{\tabcolsep}{3pt}
\renewcommand{\arraystretch}{1.15}
\begin{tabular}{@{}p{0.6cm}p{3.6cm}p{6.05cm}p{3.9cm}p{2.4cm}@{}}
\toprule
ID & Hypothesis (frozen wording) & Statistic and test (frozen) & Estimate [95\% CI] & Decision \\
\midrule
H1 & MISS costs more than every other core referent state & For each of the 5 other core referent states: paired difference cost(MISS) $-$ cost(state), macro over T6 minus that state's answer-setting cells (fixed before the freeze as the cells where a literal solver loses at least as much as with MISS: DUP on counting, SIZE $\times$2 on object size); one-sided; intersection--union & smallest of the five: vs.\ SIZE $\times$0.5 11.4 [8.8, 14.0] & supported ($p<0.001$; Holm threshold 0.0056) \\
H2 & MISS costs more than DUP & Paired difference, macro over T6 minus counting; one-sided & 23.0 [19.9, 26.0] & supported ($p<0.001$; Holm threshold 0.0063) \\
H3 & Removing half of the unnamed objects costs little (CTX-half) & TOST $\pm$3, macro over T6 & 0.3 [\ensuremath{-}1.3, 2.0] & supported ($p<0.001$; Holm threshold 0.0167) \\
H4 & Size errors cost little outside the size task & TOST $\pm$3 for SIZE $\times$0.5 and $\times$2.0, macro over T6 minus object size and absolute distance; intersection--union & $\times$0.5: 3.2 [0.8, 5.7]; $\times$2.0: 2.2 [\ensuremath{-}0.3, 4.7] & \textbf{not supported} ($p$ = 0.560; Holm threshold 0.0250) \\
H5 & 0.25\,m is tolerated; 1.0\,m is costly (LOC) & TOST $\pm$3 at 0.25\,m on the solver-selected tasks; ``costly'' test at 1.0\,m on T6; intersection--union & 0.25\,m: 0.0 [\ensuremath{-}1.7, 1.7]; 1\,m: 10.7 [8.6, 12.8] & supported ($p<0.001$; Holm threshold 0.0125) \\
H6 & A list with a missing referent scores below no list & Paired difference MISS $-$ RGB, macro over T6 minus counting; one-sided ($<$ 0) & 4.1 [0.8, 7.4] & \textbf{not supported} ($p$ = 0.993; Holm threshold 0.0500) \\
H8r & The main costs replicate on Pro & With Gemini~3.1~Pro: cost of MISS and of LOC 1.0\,m on T6, each ``costly'' (one-sided test of cost $>$ 0 rejects and estimate $\geq$ 3); intersection--union & MISS 26.1 [18.4, 34.3]; LOC 1\,m 15.0 [9.5, 21.1] & supported ($p<0.001$; Holm threshold 0.0071) \\
H10 & Gemini follows a list of another room & Paired difference X-scene $-$ RGB, macro over T6; one-sided ($<$ 0) & \ensuremath{-}6.2 [\ensuremath{-}8.9, \ensuremath{-}3.5] & supported ($p<0.001$; Holm threshold 0.0083) \\
H11 & Gemini follows a wrong list over the video & List-following rate over the direction-known cells (SIZE on size questions, MISS and DUP on counting) where the solver's answer changes; one-sided test of rate $>$ 0.5 & 88.8\% [86.5, 91.1] & supported ($p<0.001$; Holm threshold 0.0100) \\
\bottomrule
\end{tabular}
\caption{\textbf{Seven of the nine hypotheses are supported.} Hypotheses and tests as fixed before the study (Gemini~3.5~Flash-Lite; H8r: Gemini~3.1~Pro), decided by Holm's method at $\alpha$ = 0.05. H9 was dropped and H7 made exploratory before the study; H8 was replaced by H8r. Estimates are in points, except for H11 (share of answers). MISS: the asked-about object missing; DUP: duplicated; LOC: shifted; SIZE: resized; CTX-half: half of the other objects removed; X-scene: the list of another room; RGB: no list; T6: the six tasks whose questions name objects.}
\label{tab:hypotheses}
\end{table*}

\section{Compared methods}
\label{supp:methods}

\begin{figure*}[t]
\centering
\includegraphics[width=\textwidth]{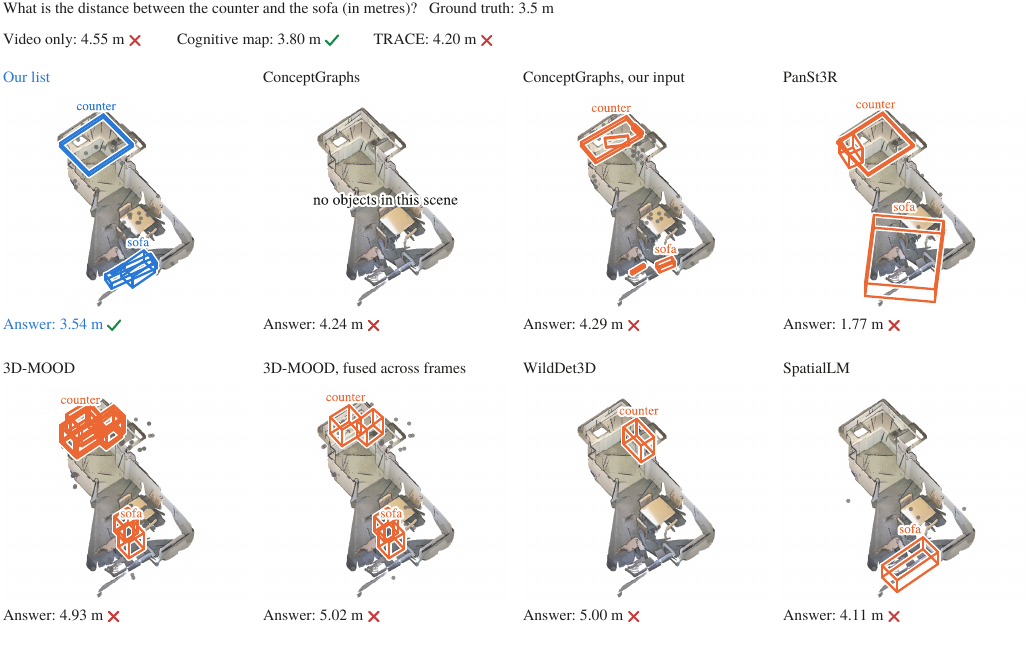}
\caption{\textbf{The lists of all compared methods for one question.} ScanNet scene0651\_02, Gemini~3.8~Flash with the same 32 frames. Each list is drawn on our reconstruction: entries named counter or sofa as boxes, all other entries as grey dots. ConceptGraphs ran on this scene but found no objects, so its list is empty. A check marks an answer within 10\% of the ground truth. We chose this example for legibility among the questions where every method ran.}
\label{fig:allmethods}
\end{figure*}

Each method's list keeps its own metric scale when we place it in our frame (Sec.~4.3). For methods that output points, our box fitter fits a box to each of their objects. The VLM sees each method's own names, with our prompt and frames. \Cref{fig:allmethods} shows the lists of all methods for one question.

\begin{figure*}[t]
\centering
\begin{minipage}{0.98\textwidth}
\scriptsize
\begin{verbatim}
The images above are frames from one video walkthrough of an indoor scene, in temporal order.
### TASK
Build a structured 3D representation of this scene. It will later be used to answer spatial
questions about the scene, which are not known yet, so include every object that someone might
ask about: furniture, appliances, fixtures, decorations and small objects. Give one entry per
physical object, counting each object once even if it appears in many frames.
Use one metric coordinate frame for the whole scene, in metres: z points up, the floor is at
z = 0, and x and y are horizontal. For each object give a short common-noun label, the center
[x, y, z] of its 3D bounding box and its size [length, width, height] along the box's own axes.
Also give the floor area of the room in square metres as room_footprint_m2.
### OUTPUT
Output only the JSON, with no other text, in exactly this format:
{"room_footprint_m2": <number>, "objects": [{"label": "<label>", "center": [x, y, z],
 "size": [l, w, h]}, ...]}
\end{verbatim}
\end{minipage}
\caption{\textbf{The request with which GPT-6 Astra builds its own list} (Sec.~4.3 of the paper). It is sent once per scene after the frames, without any question.}
\label{fig:astrabuild}
\end{figure*}

PanSt3R~\cite{zust2025panst3r} ran with its released weights on our frames. We keep its points and panoptic segments, except walls, floor and ceiling.

3D-MOOD~\cite{yang2025mood} ran with its Swin-B~\cite{liu2021swin} checkpoint on the same frames, with the ScanNet200~\cite{rozenberszki2022scannet200} class names as prompts, and we keep detections with a score of at least 0.3. We place its boxes from each frame with Pi3X's camera poses. The fused variant merges overlapping boxes with the same name (3D IoU of at least 0.25, or one centre inside the other box), which leaves about half of the boxes.

WildDet3D~\cite{huang2026wilddet3d} ran with its released code and checkpoint on the same frames. Its boxes are merged across frames with the rule of the fused 3D-MOOD variant. SpatialLM~\cite{mao2025spatiallm} (SpatialLM1.1-Qwen-0.5B) ran with its repository's settings on our Pi3X point cloud, in our frame and at our scale.

The cognitive map uses the prompt of Thinking in Space~\cite{yang2025thinking} for the objects named in the question: the VLM draws the map and then answers, in one call. TRACE~\cite{hua2026trace} uses its own prompts with our VLM and frames, so its result differs from the one it reports with Gemini~3~Pro~\cite{google2025gemini3}. GCA~\cite{chen2026gca} ran from its public code with its MMSI-Bench~\cite{yang2026mmsi} settings.

\Cref{fig:astrabuild} shows the request with which GPT-6 Astra~\cite{openai2026gpt6astra} builds its own list (Sec.~4.3).

\section{The controlled error study}
\label{supp:study}
\begin{table*}[t]
\centering
\footnotesize
\setlength{\tabcolsep}{3.5pt}
\begin{minipage}[t]{0.48\textwidth}
\centering
\begin{tabular}[t]{@{}lrccc@{}}
\toprule
Error & Qs. & VLM cost & F1@.25 & ODS \\
\midrule
\multicolumn{5}{@{}l}{\textit{Asked-about object}} \\
Missing & 1,165 & 29.8 {\scriptsize[27.3, 32.3]} & 9.1 & 4.9 \\
Duplicated & 1,165 & 10.9 {\scriptsize[8.8, 12.9]} & 6.1 & 3.0 \\
Shifted 0.1\,m & 592 & \ensuremath{-}0.8 {\scriptsize[\ensuremath{-}2.9, 1.5]} & 1.9 & 0.6 \\
Shifted 0.25\,m & 1,165 & 1.6 {\scriptsize[0.1, 3.3]} & 5.5 & 1.8 \\
Shifted 0.5\,m & 592 & 2.9 {\scriptsize[0.4, 5.3]} & 9.8 & 4.0 \\
Shifted 1\,m & 1,165 & 10.7 {\scriptsize[8.7, 12.8]} & 12.7 & 6.4 \\
Size $\times$0.8 & 592 & 5.9 {\scriptsize[3.6, 8.1]} & 0.0 & 1.1 \\
Size $\times$1.25 & 592 & 8.3 {\scriptsize[6.1, 10.6]} & 0.0 & 1.1 \\
Size $\times$0.5 & 1,165 & 18.4 {\scriptsize[16.7, 20.3]} & 13.6 & 2.0 \\
Size $\times$2 & 1,165 & 23.0 {\scriptsize[21.2, 24.8]} & 13.6 & 2.0 \\
\bottomrule
\end{tabular}
\end{minipage}\hfill
\begin{minipage}[t]{0.48\textwidth}
\centering
\begin{tabular}[t]{@{}lrccc@{}}
\toprule
Error & Qs. & VLM cost & F1@.25 & ODS \\
\midrule
\multicolumn{5}{@{}l}{\textit{Rest of the scene}} \\
Half the other objects removed & 1,008 & 0.3 {\scriptsize[\ensuremath{-}1.4, 1.9]} & 20.6 & 10.3 \\
Unrelated objects added & 572 & \ensuremath{-}1.4 {\scriptsize[\ensuremath{-}3.5, 0.8]} & 20.2 & 10.1 \\
Scale $\times$0.925 & 600 & 2.9 {\scriptsize[0.9, 5.0]} & 33.6 & 12.8 \\
Scale $\times$1.078 & 600 & \ensuremath{-}0.2 {\scriptsize[\ensuremath{-}2.7, 2.1]} & 32.5 & 13.1 \\
20\% of all removed & 572 & 12.9 {\scriptsize[9.7, 16.2]} & -- & -- \\
50\% of all removed & 594 & 27.2 {\scriptsize[24.0, 30.5]} & -- & -- \\
\midrule
\multicolumn{5}{@{}l}{\textit{Reference}} \\
No list & 1,187 & 32.2 {\scriptsize[29.4, 35.0]} & -- & -- \\
List of another room & 1,187 & 38.5 {\scriptsize[35.7, 41.3]} & -- & -- \\
\bottomrule
\end{tabular}
\end{minipage}
\caption{\textbf{The larger study: for Gemini~3.5~Flash-Lite, a missing asked-about object costs most, while a 1\,m shift costs F1 more.} VLM cost: accuracy lost against the intact ground-truth list, in points over the six tasks, with 95\% intervals. F1@.25, ODS: points lost by the same lists in F1 at 3D IoU 0.25 and in the open detection score. Qs.: the number of questions each error was run on. Errors with about 600 questions were run on the paper's 800 questions only. An error of the asked-about object needs a question that names an object.}
\label{tab:costs}
\end{table*}
\begin{table*}[t]
\centering
\footnotesize
\setlength{\tabcolsep}{5pt}
\begin{tabular}{@{}lccccccc@{}}
\toprule
 & \multicolumn{4}{c}{Detection metrics} & Object & All asked-about & Gain over \\
\cmidrule(lr){2-5}
List & F1@.25 & P@.25 & R@.25 & ODS & recall & objects found & video (Tab.~4) \\
\midrule
Ours & 0.32 & 0.27 & \textbf{0.40} & \textbf{0.59} & \textbf{0.68} & \textbf{31\%} & \textbf{5.9} \\
ConceptGraphs~\cite{gu2024conceptgraphs} & \textbf{0.33} & 0.50 & 0.28 & \textbf{0.59} & 0.40 & 5\% & 0.9 \\
PanSt3R~\cite{zust2025panst3r} & 0.24 & 0.23 & 0.28 & 0.56 & 0.64 & 26\% & 1.9 \\
3D-MOOD~\cite{yang2025mood} & 0.12 & 0.08 & 0.38 & 0.43 & 0.57 & 13\% & 1.1 \\
SpatialLM~\cite{mao2025spatiallm} & 0.29 & \textbf{0.52} & 0.22 & 0.55 & 0.30 & 6\% & 3.5 \\
\bottomrule
\end{tabular}
\caption{\textbf{Detection metrics do not tell which list helps the VLM.} The lists of Tab.~4 on the 31 VSI-Bench scenes where all five align to the ground truth. P, R: precision and recall at 3D IoU 0.25. Object recall: share of ground-truth objects found. All asked-about objects found: share of questions whose named objects are all in the list. Gain: Gemini~3.8~Flash on the 800 questions (Tab.~4). Bold: best in each column.}
\label{tab:metrics_real}
\end{table*}

The larger study of Sec.~5.1 used Gemini~3.5~Flash-Lite on all 1,587 sampled questions, with more errors and more levels of each. \Cref{tab:operators} defines every error, \cref{tab:hypotheses} gives the nine hypotheses and \cref{tab:costs} the cost of every error. Each error changes one object named in the question, and all errors of a question change the same object. For counting, it changes one object of the counted category, so a missing or duplicated object changes the true count by one. The changes are random but seeded, so they can be repeated exactly.

List-following (Sec.~5.2) is measured where we know which way an error moves the answer: a doubled box on object-size questions, and a missing or duplicated object on counting questions. We use the questions where a program that reads the list literally gives different answers for the intact and the changed list. An answer follows the list if it is closer to the program's answer for the changed list than to its answer for the intact list. H11 in \cref{tab:hypotheses} is the same measure for Gemini~3.5~Flash-Lite.

F1 at 3D IoU 0.25 (Sec.~5.2) pairs list and ground-truth boxes one to one within a category, highest IoU first. We compute the 3D IoU exactly: for gravity-aligned boxes, it is the overlap of the two footprints times the overlap of their heights. The open detection score of 3D-MOOD~\cite{yang2025mood} is $\text{ODS} = [3\,\text{AP}_\text{dist} + (1-\text{TE}) + (1-\text{SE}) + (1-\text{OE})]/6$, with F1 in place of $\text{AP}_\text{dist}$. It pairs boxes by the distance of their centres, at 0.5 to 1.0 times half the diagonal of the ground-truth box. TE is that distance relative to the threshold, SE is one minus the IoU after aligning centre and orientation, and OE is the angle between the boxes divided by $\pi$.

\begin{table}[!htb]
\centering
\footnotesize
\setlength{\tabcolsep}{3pt}
\begin{tabular}{@{}lc@{}}
\toprule
 & Points [95\% interval] \\
\midrule
Gap of the list to the ground-truth list & 24.1 [19.4, 29.0] \\
Gain from all five fixes together & 13.4 [8.2, 18.6] \\
\midrule
\multicolumn{2}{@{}l}{\textit{Lost when one fix is left out}} \\
\quad Missed objects added & 8.5 [4.3, 12.6] \\
\quad Boxes snapped to ground truth & 4.6 [\ensuremath{-}0.2, 9.1] \\
\quad Duplicates removed & 1.9 [\ensuremath{-}2.0, 5.9] \\
\quad Background objects removed & 1.2 [\ensuremath{-}2.6, 5.2] \\
\quad Labels corrected & 0.0 [\ensuremath{-}4.1, 4.1] \\
\bottomrule
\end{tabular}
\caption{\textbf{Fixing real lists.} Only the restored missing objects are needed for a significant gain. }%
\label{tab:oracle}
\end{table}

\Cref{tab:metrics_real,tab:oracle} test these findings on real lists (Sec.~5.2). \Cref{tab:metrics_real} gives the detection metrics of five lists of Tab.~4. The oracle fixes of \cref{tab:oracle} start from an earlier version of our list. With the ground truth, they add the missed objects, remove duplicates, correct names, snap boxes to the ground truth and remove background objects.

\end{document}